\documentclass[a4paper,fleqn]{cas-sc}

\usepackage[authoryear]{natbib}
\usepackage{amsmath,amssymb,bm}
\usepackage{booktabs,array}
\usepackage{siunitx}
\newcommand{\Frm}{Frenet}

\begin{document}
\let\WriteBookmarks\relax
\def\floatpagepagefraction{1}
\def\textpagefraction{.001}
\shorttitle{Energy-Aware Path Following: Comparative Analysis of Reinforcement Learning and NMPC for Electric Vehicles}
\shortauthors{M. Sabaa and M. Emam}

\title [mode = title]{Closing the Regenerative-Energy Gap in Learned Path
  Following: Preview-Augmented Reinforcement Learning versus NMPC for
  Electric Vehicles in the \Frm{} Frame}

\author[1]{Mohamed Sabaa}[orcid=0009-0007-8963-7460]
\ead{444307237@nu.edu.sa}
\credit{Conceptualization, Literature Research, Methodology, Software,
  Formal Analysis, Investigation, Visualization, Writing -- Original Draft.}
\affiliation[1]{organization={Computer Science Department,
      Najran University},
  city={Najran},
  country={Saudi Arabia}}

\author[2]{Mostafa Emam}[orcid=0000-0003-4942-1183]
\cormark[1]
\ead{mostafa.emam@unibw.de}

\credit{Conceptualization, Software, Supervision, Validation,
  Writing -- Review and Editing}
\affiliation[2]{organization={Institute of Applied Mathematics and
      Scientific Computing,
      University of the Bundeswehr Munich},
  city={Neubiberg},
  postcode={85579},
  country={Germany}}

\cortext[cor1]{Corresponding author}

\begin{abstract}
  Path-following control strategies typically follow the bi-objective optimization dilemma: minimizing deviations from a reference path while maintaining smooth speed profiles. The latter objective is especially relevant for Electric Vehicles (EVs), since their limited driving range can be extended by recovering energy through regenerative braking, a feature that has not yet been sufficiently studied in the literature. In this work, we perform a comparative analysis of four controllers under one common Frenet frame-based kinematic vehicle model, utilizing a validated energy model (VT-CPEM) with explicit regenerative braking. Herein, we implement the following controllers: Nonlinear Model Predictive Control (NMPC), Proximal Policy Optimization (PPO), gain-scheduled Ackermann state-feedback baseline (PID-SF), and a Stanley geometric baseline. To satisfy real-time requirements, we implement the NMPC using JIT-compiled CasADi. Moreover, we train the PPO using traditional straight and S-curve tracks, after which we successfully transfer the unmodified policy to unseen tracks, including: an ISO~3888-1 lane-change, a chicane, randomly-generated parameterized-splines, and a $\pm3^\circ$ graded road. In addition, the policy transfers to a dynamic single-track vehicle model with linear tires, zero-shot with an acceptable initial performance, which was optimized after brief fine-tuning. Thereby, we demonstrate that our PPO is readily transferable to more comprehensive vehicle models. We conclude with a performance analysis of developed controllers and discuss ideas for future work.

\end{abstract}

\begin{highlights}
  \item Benchmarking the tracking--regeneration trade-off under a unified EV energy model.
  \item Preview observation $+$ energy-based reward close $76$\% of NMPC's energy gap.
  \item Same change improves tracking: RMSE$_d$ $0.459\!\to\!0.314~(\si{m})$.
  \item $0/30$ failures over LHS initial conditions; Wilcoxon $p<0.0001$, $d=0.826$.
  \item Policy transfers in simulation experiments to unseen tracks, road grade, and a tire-dynamics plant.
\end{highlights}


\maketitle

\section{Introduction}
\label{sec:intro}

Among its advantages, Autonomous Driving (AD) is pursued because an estimated 94\% of crashes are attributable to driver (i.e., \emph{human}) error, and as it promises to extend personal mobility to individuals unable to drive themselves~\cite{Paden2016}. Electric vehicles (EVs) are increasingly becoming the platform of choice for this transition, yet they remain constrained by limited driving range, which makes accurate, real-time estimation and management of energy consumption during travel a first-order design requirement rather than an afterthought \cite{Ding2024}.

The decision-making pipeline of a self-driving vehicle is commonly decomposed into \emph{route planning}, \emph{behavioral decision-making}, \emph{motion planning}, and \emph{local feedback control}~\cite{Paden2016,Wu2024Survey,Zhao2024Survey}. This paper addresses the last layer. Herein, a reference path and a path-parameterized reference speed trajectory are assumed to be provided by an upstream planner, where the developed controllers are tasked with computing steering-rate and longitudinal-acceleration/deceleration commands in real-time, with the objective of minimizing tracking errors while adhering to actuator bounds. To meet strict real-time requirements, path-following controllers typically decouple these dynamics, addressing longitudinal and lateral tracking independently or sequentially. However, due to the nonholonomic nature of EVs, the resulting control commands are suboptimal~\cite{Paden2016,Liu2021}. Moreover, the performance evaluation benchmarks predominately focus on tracking errors without reporting energy metrics, excluding a fundamental criterion for the deployment of EVs~\cite{Liu2021,Lee2023Compare}. This work intends to address this research gap by: (i) accommodating the path-following problem with coupled dynamics, (ii) satisfying strict real-time requirements, and (iii) providing a comprehensive analysis covering both tracking errors and energy-consumption metrics. Thereby, we establish a theoretical foundation and provide practical benchmarking results for future studies investigating real-time, energy-aware path-following controllers with coupled dynamical models.

\section{Literature Review}
\label{sec:related}

Nonlinear model predictive control (NMPC) has emerged as a central framework for path-following in AD applications, owing to its explicit handling of multi-variable vehicle dynamics, constraints, and performance objectives within a receding-horizon optimization setting \cite{Rawlings2017}. Recent works highlight the continued relevance of MPC-based controllers, in particular due to its ability to integrate trajectory tracking, constraint satisfaction, and robustness considerations into a unified formulation \cite{Stano2023,Zhao2024Survey}. For example, \cite{Belkebir2026} addressed real-time path tracking by integrating path-following and curvature-aware velocity tracking objectives into a unified NMPC cost function, solving the resulting optimization problem with CasADi/IPOPT for real-time feasibility. Beyond nominal tracking, other studies focused on operation near the limit of handling, where accurate prediction of tire and vehicle dynamics is critical. For instance, Fu et al. proposed an NMPC strategy with stable limit handling, while Domina and Tihanyi developed an LTV-MPC formulation for path-following at the handling limits \cite{Fu2022,Domina2022}. Reiter et al. demonstrated the importance of coordinate and model representations, showing that Frenet-Cartesian formulations can improve NMPC-based obstacle avoidance and tracking performance in AD contexts \cite{Reiter2023}. Consequently, these contributions emphasize that NMPC is particularly well-suited for safe path-following when both geometric accuracy and dynamic feasibility are required \cite{Rawlings2017,Stano2023}.

Contrary to traditional methods, Reinforcement learning (RL) provides a data-driven alternative for vehicle control, offering a flexible approach to situations where hand-crafted control laws are costly to tune across varying driving conditions. Early works demonstrated that RL and deep learning can be effectively utilized for lateral control in AD applications, motivating further research into learning-based path-following controllers \cite{Li2018}, with subsequent studies focusing on policy learning, employing deep deterministic approaches and reward-shaping strategies tailored to autonomous racing and trajectory tracking \cite{Hess2021,Evans2021}. More recently, RL has been combined with model-based control, for instance through MPC demonstrations or stochastic MPC structures to enhance stability and sample efficiency \cite{Wu2025,Zarrouki2024}. In addition, Frenet-frame formulations have been adopted to simplify trajectory learning and improve geometric interpretability in autonomous driving tasks \cite{Yoon2024}. Survey studies further highlight the increasing relevance of RL-based methods for AD behavior planning; however, these methods remain heavily reliant on careful design of reward functions, adequate choice of training data, and effective integration with safety-oriented fallback conventional controllers \cite{Wu2024Survey}.

Curriculum Learning (CL) naturally complements RL by structuring the training process from simpler to more complex driving tasks, thereby improving convergence and policy robustness \cite{Bengio2009}. In AD applications, CL has been implemented in staged training regimes for learning lane-keeping and maneuvering under increasingly challenging conditions \cite{Anzalone2021}. Furthermore, B\'{e}csi proposed RRT-guided experience generation to bias exploration toward informative trajectories for autonomous lane keeping, interpreting this as a curriculum-like mechanism for accelerating RL training \cite{Becsi2024}. These works indicate that curriculum design is particularly valuable in RL-based vehicle control, where direct end-to-end learning from scratch is often inefficient and unstable.

With the increasing adoption of EVs, researchers began to investigate energy-aware vehicle control strategies, since electrified powertrains inherently integrate energy efficiency as an essential component of motion planning and control. Fiori et al. developed and validated a power-based energy consumption model, establishing a foundation for incorporating energy terms into control and planning objectives \cite{Fiori2016}. More recent studies have utilized this approach to refine control actions, such as optimal regenerative braking strategies that consider braking intention and enhance energy recuperation \cite{Tang2024}. Furthermore, NMPC-based approaches have been extended to accommodate safety and stability requirements while adhering to the dynamics of EVs \cite{Liu2024}. At the planning level, energy-efficient hybrid MPC and trajectory optimization, incorporating energy-model-aware design, indicate that trajectory generation algorithms can achieve balanced behavior between travel efficiency, drivability, and energy consumption \cite{Ding2024, Tian2025}. Collectively, these developments highlight that energy awareness is becoming an essential design criterion in AD control, particularly for EVs.

As previously mentioned, path-following controllers typically do not explicitly include an energy model, limiting the driving range and negatively impacting the adoptability of EVs. Moreover, energy-aware controllers generally prioritize longitudinal or velocity control, thereby limiting an accurate representation of the consumed energy due to suboptimal lateral behavior. A notable exception is the work of \cite{Tian2025}, which addresses energy-aware trajectory optimization by generating feasible quintic candidate trajectories within the Frenet frame and, subsequently, refining them through an online nonlinear programming (NLP) step to enhance fuel efficiency. However, since the approach relies on candidate generation, it offers no guarantees regarding optimality nor robustness. Furthermore, the study does not report solution times, raising questions regarding its real-time applicability.

\begin{table}[pos=htb]
  \centering
  \caption{Comparative overview of representative control and planning methods discussed in this section.}
  \label{tab:comparative_control_methods}
  \scriptsize
  \renewcommand{\arraystretch}{1.2}
  \begin{tabular}{@{}
    >{\centering\arraybackslash}p{2.5cm}
    >{\centering\arraybackslash}p{3.4cm}
    >{\centering\arraybackslash}p{2.5cm}
    >{\centering\arraybackslash}p{1.6cm}
    >{\centering\arraybackslash}p{3.5cm}
    @{}}
    \hline
    \textbf{Research and Scope}                                          & \textbf{Vehicle Model}                                                          & \textbf{Control Strategy}                                   & \textbf{Energy Model}          & \textbf{Experiments}                                                                     \\
    \hline
    \cite{Fu2022}: Path tracking at handling limits                      & Nonlinear vehicle dynamics with tire-force, load-transfer, and adhesion effects & NMPC                                                        & -                              & HIL and simulation using CarSim.                                                         \\
    \cite{Domina2022}: Automated path following near handling limits     & Linear time-varying vehicle model considering steering dynamics                 & LTV-MPC                                                     & -                              & Simulation-based using MATLAB.                                                           \\
    \cite{Reiter2023}: Obstacle avoidance and path following             & Kinematic vehicle model in combined Cartesian/Frenet coordinate representation  & NMPC                                                        & -                              & Simulation-based using ACADOS.                                                           \\
    \cite{Belkebir2026} Curvature-aware path tracking with embedded NMPC & Augmented kinematic model in Frenet frame                                       & NMPC with curvature-aware speed and Lyapunov terminal cost  & -                              & Simulation using CARLA and embedded timing with CasADi on Raspberry Pi 5 and Xavier AGX. \\
    \cite{Li2018}: Vision-based lateral control                          & TORCS / VTORCS simulator with a perception-to-control pipeline                  & RL + deep learning perception                               & -                              & Simulator-based training and evaluation using VTORCS; compared against LQR, MPC.         \\

    \cite{Hess2021}: Longitudinal and lateral path following             & Custom kinematic bicycle model                                                  & DDPG                                                        & -                              & Simulation-based using PyTorch.                                                          \\
    \cite{Evans2021}:  Autonomous racing local planning                  & Kinematic vehicle model (scaled F1/10th)                                        & RL with reward design                                       & -                              & Simulation-based using OpenAI-Gym.                                                       \\
    \cite{Wu2025}: Lateral control                                       & Time-invariant continuous dynamic linear model                                  & DRL with MPC-PID demonstration                              & -                              & Simulator-based using CARLA.                                                             \\
    \cite{Zarrouki2024}: Motion control under uncertainty                & Dynamic nonlinear single-track model                                            & RL-driven adaptive Stochastic NMPC                          & -                              & Simulation-based using TUM-CONTROL framework.                                            \\
    \cite{Yoon2024}:  Trajectory learning in the Frenet frame            & Frenet-frame road geometry for curved/urban scenarios                           & RL-based trajectory learning                                & -                              & Simulation-based using Gazebo and ROS.                                                   \\
    \cite{Anzalone2021}: Autonomous driving in simulation                & CARLA simulator; end-to-end driving policy                                      & Reinforced curriculum learning + DRL                        & -                              & Simulator-based using CARLA.                                                             \\
    \cite{Fiori2016}: EV energy-consumption modeling                     & Power-based EV energy consumption model                                         & -                                                           & Same as vehicle model          & Evaluated on actual EVs, including Nissan Leaf and Tesla model S.                        \\
    \cite{Tang2024}: Regenerative braking control                        & EV braking, load, and battery-SoC-dependent recovery model                      & Deterministic (flowchart-based) analytical braking strategy & Same as vehicle model          & HIL and simulation using dSPACE ASM Suite.                                               \\
    \cite{Ding2024}: Energy-efficient velocity planning                  & Kinematic vehicle model with motion forces and passive energy recovery          & Hybrid controller using Dynamic- and Quadratic-Programming  & Same as vehicle model          & Simulation-based using Matlab, Carsim, and Prescan.                                      \\
    \cite{Tian2025}: Energy-aware trajectory optimization                & Frenet polynomial trajectories with differentiable fuel model                   & EMATO: online NLP/BVP over quintic Frenet candidates        & Differentiable fuel-rate model & Simulation-based on custom sedan and truck models.                                       \\
    \hline
  \end{tabular}
\end{table}

The remainder of this work is structured as follows:
\begin{itemize}
  \item We start with presenting our fundamental vehicle model; a coupled kinematic vehicle model in the Frenet frame, providing a compact yet adequate basis for longitudinal--lateral path-following control.
  \item Using this model, we develop and compare four control strategies: (i) an NMPC controller implemented with JIT acceleration via CasADi and tuned using Latin Hypercube Sampling (LHS), (ii) an RL-PPO controller with preview information and an energy-aware reward designed through five-phase CL, (iii) a PID-SF controller based on Ackermann state feedback and tuned with pole placement, and (iv) a Stanley-geometric baseline with a proportional inner loop and a longitudinal PI law shared with PID-SF.
  \item Afterwards, we perform a dedicated comparative study of the RL-PPO reward design, including the five CL phases, to quantify its influence on learning and closed-loop behavior. Similarly, we discuss the policy transferability and undertaken measures to enhance robustness against noise and unmodelled disturbances.
  \item We conduct an energy analysis for our developed controllers, reporting the overall energy consumption and the amount of successfully recovered energy through regenerative braking.
  \item For a more comprehensive evaluation, we simulate our controllers on a dynamic single-track plant and across standardized tracks, including an additional LQR baseline for reference.
  \item Finally, we summarize our findings and present our ideas for future work.
\end{itemize}

\section{System Formulation}
\label{sec:system}
Having specified the scope of this work, we now construct the shared mathematical framework, upon which our controllers operate. Herein, three components interact by design: the kinematic model defines vehicle motion in the Frenet frame, the velocity profile imposes a curvature-aware speed demand, and the energy model converts that demand into a scalar quantity, which is comparable across the different controllers.

\subsection{Kinematic Vehicle Model}
\label{ssec:kinematics}
Compared to the Cartesian representation, the Frenet coordinate system~\cite{Werling2010} describes vehicle motion relative to a reference path, decomposing the position into traversed arclength~$s$ along the path and signed lateral deviation~$d$ from it, taken positive to the \emph{left} of the path. More formally, let $r(s): [0,s_{\max}] \rightarrow \mathbb{R}^2$ be a reference path parameterized by arclength, e.g., using a natural cubic spline. The vehicle state is defined as $\bm{x}=[s,\,d,\,\alpha,\,\delta,\,v]^{\top}$, with the arclength $s~(\si{m})$, lateral deviation $d~(\si{m})$, heading error $\alpha~(\si{rad})$, front-wheel steering angle $\delta~(\si{rad})$, and longitudinal speed $v~(\si{ms}^{-1})$ with respect to the travel direction~\cite{Reiter2023}. The reference path is characterized by its curvature $\kappa(s)~(\si{m}^{-1})$, and the vehicle is actuated through the controls $\bm{u}=[u_1,\,u_2]^T$, where $u_1 = \dot{\delta}~(\si{rad\,s}^{-1})$ and $u_2 = \dot{v}~(\si{ms}^{-2})$ denote the steering-rate and acceleration commands, respectively. The model dynamics are given by:
\begin{align}
  \dot{s}      & = \frac{v\cos\alpha}{1-d\kappa(s)}, \label{eq:sdot}         \\
  \dot{d}      & = v\sin\alpha, \label{eq:ddot}                              \\
  \dot{\alpha} & = \frac{v\tan\delta}{L_f}-\kappa(s)\dot{s}, \label{eq:adot} \\
  \dot{\delta} & = u_1,\qquad \dot{v}=u_2. \label{eq:ddelta}
\end{align}
where $L_f~(\si{m})$ denotes the wheelbase. For real- and discrete-time systems, \eqref{eq:sdot}--\eqref{eq:ddelta} are typically discretized and computed using fourth-order Runge--Kutta integration, e.g., with a sampling time of $T_s=0.1~(s)$. Moreover, the model dynamics are only valid when the projection unto the reference path is unique and the denominator in \eqref{eq:sdot} remains separated from zero. These conditions are explicitly verified and guaranteed during the construction phase of reference paths for training and evaluation purposes, where we enforce:
\begin{equation}
  1 - d \kappa(s) \ge \epsilon_d > 0,
\end{equation}
with the numerical safety guard $\epsilon_d = 10^{-6}$, which is typically sufficient for urban scenarios. For example, the extreme corridor failure boundary $\max|d| = 4.9~(\si{m})$ on the benchmark S-curve with $\kappa_{\max} = 0.1101~(\si{m}^{-1})$ yields a maximum value of $\max\limits_{s\in[0,s_{\max}]} |d\kappa(s)| \approx 0.54 < 1.0$, confirming that no division-by-zero singularity occurs prior to episode termination. For coordinate-transformation, we implement a Newton-method-based minimization function to identify and uniquely transform between Cartesian and Frenet coordinates. Furthermore, to preserve the validity of the no-slip kinematic assumption, we define a curvature-aware upper bound on the vehicle's longitudinal velocity derived from:
\begin{equation}
  a_{\mathrm{lat}} = \frac{v^2|\tan\delta|}{L_f} \le a_{\mathrm{lat},\max},
  \label{eq:alat_check}
\end{equation}
with the maximum lateral acceleration $a_{\mathrm{lat},\max} = 2.5~(\si{ms}^{-2})$. To simplify the control problem, \eqref{eq:alat_check} is enforced as a minimization objective rather than a hard constraint; therefore, we include $a_{\mathrm{lat}}$ as a reporting metric across our training and validation scenarios to prove that no constraint violations occur.

\subsection{Reference Velocity Generation}
\label{ssec:vref}

For the given reference path, we define the curvature-aware reference speed as:
\begin{equation}
  v_{\mathrm{ref}}(s)=\min\!\left( v_{\max},\; \sqrt{\frac{a_{\mathrm{lat},\max}}{|\kappa(s)|+\epsilon_\kappa}},\; \sqrt{2\,a_{\mathrm{lng},\max}\,\max(s_{\max}-s, 0)} \right), \label{eq:vref}
\end{equation}
augmenting the work of \cite{Belkebir2026} and employing a moving-average filter with a three-step window to suppress spline-derivative noise. Here, $v_{\mathrm{ref}}^2 |\kappa(s)| \le a_{\mathrm{lat},\max}$ ensures that $a_{\mathrm{lat}}$ does not exceed the maximum allowed lateral acceleration $a_{\mathrm{lat},\max}$ when operating at the dynamic limits; the regularization term $\epsilon_\kappa~(\si{m})$ prevents division by zero on straight segments. Furthermore, we adopt a smooth terminal stopping profile with $\sqrt{2a_{\mathrm{lng}}\max(s_{\max}-s,0)}$, which emphasizes the comfort deceleration limit $a_{\mathrm{lng},\max}~(\si{ms}^{-2})$ until reaching standstill at the end of path, i.e., at $s=s_{\max}$. Note that the terminal ramp contributes the largest proportion of sustained negative power recovered by the energy model, thereby serving as the dominant source of regenerative energy fractions reported in subsequent sections of this study.

\begin{figure}[pos=htb]
  \centering
  \includegraphics[width=0.5\linewidth]{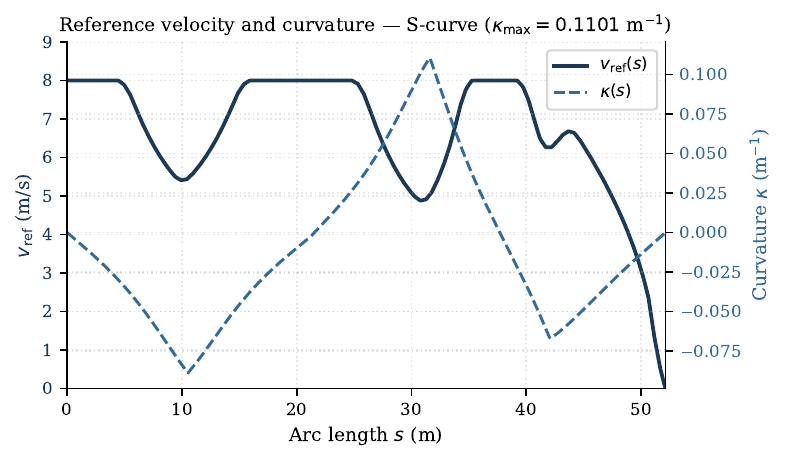}
  \caption{Reference velocity $v_{\mathrm{ref}}(s)$ and curvature
    $\kappa(s)$ on an S-curve. The terminal deceleration ramp drives regenerative energy recovery.}
  \label{fig:vref}
\end{figure}

\subsection{Electric Energy Model}
\label{ssec:energy_model}
We utilize VT-CPEM~\cite{Fiori2016} for determining the required traction power and recoverable energy fraction. Here, the road-load force is given by:
\begin{equation}
  F(v,a)=ma+\frac{1}{2}\rho C_D A_f v^2 + m g C_r\cos\theta_{\mathrm{road}}+ m g \sin\theta_{\mathrm{road}}, \label{eq:force}
\end{equation}
where the parameters are delineated in Table~\ref{tab:params}, yielding the required wheel power $P_w = F v$. A consistent battery-side power boundary can be written as:
\begin{equation}
  P_{batt} = \begin{cases}
    P_{aux} + \eta_t P_w, & P_w \ge 0, \\
    P_{aux} + \eta_r P_w, & P_w < 0,   \\
  \end{cases}
\end{equation}
with the traction (drivetrain) efficiency $\eta_t$ and energy-recovery efficiency $\eta_r$. Notably, \cite{Fiori2016} define $\eta_r$ as a deceleration-dependent feature, which we assume here to be constant to disentangle dependencies and reduce the computational burden on the real-time NMPC. We amend this through a sensitivity analysis over a comprehensive range of $\eta_r$ values, which we later explain in Section~\ref{ssec:eta}.

The auxiliary power $P_{aux}$ represents energy consumed by vehicle electronics, infotainment systems, and internal ECUs. This variable is significantly affected by operating conditions, such as temperature, battery state, and currently active electronic components, rendering it challenging to incorporate into simulation studies \cite{Kropiwnicki2023}. In our case, we primarily focus on different controller energy requirements; specifically, $P_{aux}$ reduces to the energy consumed by a simulative ECU computing the control commands. Using the discretization sampling time $T_s$, we define the energy quantities as:
\begin{equation}
  E_t = \eta_t\max(P_w, 0)~T_s, \qquad E_r = \eta_r \max(-P_w, 0)~T_s, \qquad E_{net}= E_t + E_{ECU} - E_r,
\end{equation}
with the traction energy $E_t$, the recovered energy $E_r$, the controller energy requirement $E_{ECU} = \sum\limits_{k} P_{ECU} t_{exe,k}$ summed over $k$ solving steps, and the total consumed energy $E_{\mathrm{net}}$. Since the energy split depends strictly on the sign of mechanical wheel power $P_w$ rather than vehicle acceleration, gentle decelerations, where traction force remains positive due to aerodynamic drag and rolling resistance, are correctly assigned to $E_t$ and excluded from regenerative recovery $E_r$. Note that $E_{ECU}$ is typically negligible for the quasi-instantaneous controllers (PPO and PID-based), yet it is comparatively higher for the optimization-based NMPC. Nevertheless, we retain it as a fairness safeguard rather than a discriminator.

\section{Controller Design}
\label{sec:ctrl}

\begin{table}[pos=htb]
  \renewcommand{\arraystretch}{1.10}
  \caption{Complete Parameter List.}
  \label{tab:params}
  \centering
  \scriptsize
  \begin{tabular}{lll}
    \toprule
    Symbol                                  & Value                                 & Description                                             \\
    \midrule
    \multicolumn{3}{l}{\textit{Vehicle (adopted from VT-CPEM~\cite{Fiori2016})}}                                                              \\
    $m$                                     & $1500~(\si{kg})$                      & Mass                                                    \\
    $L_f$                                   & $4.0~(\si{m})$                        & Wheelbase                                               \\
    $C_D$                                   & $0.30$                                & Drag coefficient                                        \\
    $A_f$                                   & $2.2~(\si{m}^2)$                      & Frontal area                                            \\
    $C_r$                                   & $0.01$                                & Rolling resistance                                      \\
    $\rho$                                  & $1.225~(\si{kg\,m}^{-3})$             & Air density                                             \\
    $\eta_r$                                & $0.65$                                & Regen efficiency (constant, cf. Section~\ref{ssec:eta}) \\
    $P_{\mathrm{ECU}}$                      & $5~(\si{W})$                          & Embedded ECU power draw (estimated)                      \\
    \midrule
    \multicolumn{3}{l}{\textit{Dynamic model (Section~\ref{sec:dynamic})}}                                                                    \\
    $I_{zz}$                                & $3000~(\si{kg\,m}^2)$                 & Yaw inertia                                             \\
    $l_f,l_r$                               & $2.0, 2.0~(\si{m})$                   & Axle distances from CoG                                 \\
    $C_f,C_r^{\mathrm{tire}}$               & $80~(\si{kN\,rad}^{-1})$              & Cornering stiffness~\cite{Rajamani2012}                 \\
    \midrule
    \multicolumn{3}{l}{\textit{State and input bounds}}                                                                                       \\
    $d$                                     & $[-5,5]~(\si{m})$                     & Lateral deviation                                       \\
    $\alpha$                                & $[-\pi,\pi]~(\si{rad})$               & Heading error                                           \\
    $\delta$                                & $[-0.6,0.6]~(\si{rad})$               & Steering angle                                          \\
    $v$                                     & $[0,8]~(\si{ms}^{-1})$                & Speed                                                   \\
    $u_1$                                   & $[-0.4,0.4]~(\si{rad\,s}^{-1})$       & Steering rate                                           \\
    $u_2$                                   & $[-2.5,2.5]~(\si{ms}^{-2})$           & Acceleration                                            \\
    \midrule
    \multicolumn{3}{l}{\textit{Reference velocity profile \eqref{eq:vref}}}                                                                   \\
    $v_{\max}$                              & $8.0~(\si{ms}^{-1})$                  & Maximum speed                                           \\
    $a_{\mathrm{lat}}$                      & $2.5~(\si{ms}^{-2})$                  & Lateral acceleration limit                              \\
    $a_{\mathrm{lng}}$                      & $3.0~(\si{ms}^{-2})$                  & Longitudinal deceleration limit                         \\
    $T_s$                                   & $0.1~(\si{s})$                        & Discretization (sampling) period                        \\
    \midrule
    \multicolumn{3}{l}{\textit{NMPC}}                                                                                                         \\
    $N_h$                                   & $50$                                  & Number of horizon (prediction) points                   \\
    $w_d$, $w_\alpha$, $w_v$                & $10, 100, 1$                          & Lateral, heading, speed weights                         \\
    $w_{u_1}$, $w_{u_2}$                    & $1, 1$                                & Steering and acceleration weights                       \\
    \midrule
    \multicolumn{3}{l}{Gain-Scheduled Full-State Feedback \textit{PID-SF}}                                                                    \\
    $K_d^0$, $K_\alpha^0$, $K_\delta$       & $3.0, 19.0, 9.0$                      & Lateral gains computed at $v_0=5~(\si{ms}^{-1})$        \\
    $K_p^{\mathrm{lon}},K_i^{\mathrm{lon}}$ & $1.0, 0.02$                           & Longitudinal gains                                      \\
    \midrule
    \multicolumn{3}{l}{\textit{Stanley~\cite{Hoffmann2007}}}                                                                                  \\
    $k_e$, $k_{\mathrm{soft}}$, $K_\delta$  & $1.0, 1.0, 5.0$ & Cross-track, softening, inner loop                      \\
    \midrule
    \multicolumn{3}{l}{\textit{PPO~\cite{Raffin2021}}}                                                                                        \\
    $\gamma$, $\lambda_{\mathrm{GAE}}$      & $0.99, 0.95$                          & Discount, GAE                                           \\
    $\varepsilon_{\mathrm{clip}}$, batch    & $0.2, 256$                            & Clip range, mini-batch                                  \\
    network, $\eta$                         & $[256,256], 3{\times}10^{-4}$         & MLP layers, Adam LR                                     \\
    $L_s^i$                                 & $5, 15, 30~(\si{m})$                  & Preview lookahead distances, cf. Section~\ref{ssec:ppo} \\
    seeds                                   & \{42,\dots,51\} & Ten training seeds           \\
    \bottomrule
  \end{tabular}
\end{table}

\subsection{NMPC}
\label{ssec:nmpc}

Our first strategy represents the optimal control approach, which is dedicated to identifying the \textit{best} control inputs (as determined by a predefined objective function) for a feasible problem. Consequently, this approach guarantees an optimal solution, albeit often at the cost of increased computational resources. At each step $k$, the solver tries to find the controls that minimize the objective function:
\begin{subequations}
  \begin{align}
    \min_{\bm{u}}\; &
    \sum_{j=0}^{N_h-1}\!\left[
      w_d\!\left(\!\frac{d}{d_{\max}}\!\right)^{\!2}
      \!+w_\alpha\!\left(\!\frac{\alpha}{\pi}\!\right)^{\!2}
      \!+w_v\!\left(\!\frac{v-v_{\mathrm{ref}}}{v_{\max}}\!\right)^{\!2}
      \!+\!\sum_{i=1}^{2}\! w_{u_i}\!\left(\!\frac{u_i}{u_{i,\max}}\!\right)^{\!2}
      \right],
    \label{eq:nmpc_cost} \\
    \mathrm{s.t.}\; &
    \bm{x}_{k+j+1}=f(\bm{x}_{k+j},\bm{u}_{k+j}),\;\;
    \bm{x}\in\mathcal{X},\;\bm{u}\in\mathcal{U}.
  \end{align}
\end{subequations}

Each term is normalized by the square of its own bound, so that the five cost
components are dimensionless and comparably scaled; in other words, we utilize the effective weights $w_i/x_{i,\max}^2$. Here, $\mathcal{X},\mathcal{U}$ are the bound sets of Table~\ref{tab:params} and $f(\cdot,\cdot)$ is the RK4 discretization of \eqref{eq:sdot}--\eqref{eq:ddelta}. We employ a horizon of $N_h=50$ steps, yielding a time of $5~(\si{s})$, and solve the problem with IPOPT~\cite{Wachter2006} with a maximum number of iterations $\texttt{max\_iter}=20$ to respect the real-time requirements. Critically, the closed-loop plant is the same nonlinear \Frm{} model used by every other controller, such that no controller possesses a model advantage.

\subsubsection{Warm-Starting}
\label{ssec:warmstart}
To further accelerate the NMPC solving time, we initialize IPOPT with the shifted solution from the previous solution step, thereby reducing the required iterations and solve time, on average, from $124.4~(\si{ms})$ to $90.6~(\si{ms})$, i.e., a $1.37\times$ speedup. Note that warm-starting biases the solver towards the previous plan, which may be undesired in the case of highly-dynamic environments ~\cite{Rawlings2017}; however, it is practical and justified in our application.

\subsection{Gain-Scheduled Full-State Feedback (PID-SF)}
\label{ssec:pid}
The second approach follows traditional error-correction control techniques \cite{Yang2020}, where we split the control law into two independent longitudinal speed and lateral deviation stabilizers. For lateral control, we linearize the nonlinear system dynamics \eqref{eq:ddot}--\eqref{eq:ddelta} at a constant speed $v_0=5~(\si{ms}^{-1})$, such that, for small values of $\alpha, \delta$ on a locally straight path, the lateral error dynamics become:
\begin{equation}
  \dot{d} = v \alpha, \qquad \dot{\alpha} = \frac{v}{L_f}\delta, \qquad \dot{\delta} = u_1, \label{eq:lin_lat_dyn}
\end{equation}
with the control law $u_1 = - K_d d - K_\alpha \alpha - K_\delta \delta$. Consequently, we get the characteristic polynomial:
\begin{equation}
  p(\lambda) = \lambda^{3}+K_\delta\lambda^{2}+\frac{v}{L_f} K_\alpha \lambda +\frac{v^2}{L_f}K_d,
  \label{eq:charpoly}
\end{equation}
which, using pole-placement methods to identify the poles $\{-1.5,-2.5,-5.0\}$, yields:
\begin{equation}
  p_d(\lambda) = \lambda^{3} + 9\lambda^{2} + 23.75\lambda + 18.75.
  \label{eq:charpoly_num}
\end{equation}
For $v_0=5~(\si{ms}^{-1})$ and $L_f=4.0~(\si{m})$, we get the lateral gain $K_d=3.0$, heading gain $K_\alpha=19.0$, and steering gain $K_\delta=9.0$. Note that the gains must be adequately scaled to preserve the constant-speed linearization coefficients of \eqref{eq:charpoly}, in which we adaptively compute:
\begin{equation}
  K_d(v) = K_d^0 \left(\frac{v_0}{v_s}\right)^2, \qquad K_\alpha(v) = K_\alpha^0 \frac{v_0}{v_s}, \qquad v_s = \max(v, v_{\min}),
\end{equation}
where the clamp $v_{\min}=0.5~(\si{ms}^{-1})$ avoids unboundedly-increasing controls as $v\to0$. Finally, we improve the handling at curved paths (cf. \cite{Paden2016}) by defining the kinematic steering reference $\delta_{\mathrm{ref}}(s) = \arctan(L_f \kappa(s))$, with:
\[\dot{\delta}_{\mathrm{ref}}=\frac{L_f\dot{\kappa}(s)}
  {1+\left(L_f\kappa(s)\right)^2}, \]
and, thereby, amend the control law to become:
\begin{equation}
  u_1 = - K_d(v)\,d - K_\alpha(v)\,\alpha - K_\delta\,\delta + K_{ff}\,\kappa(s)\,L_f.
  \label{eq:pidsf}
\end{equation}

We record that the closed-loop matrix of the constant-speed linearized model is Hurwitz-stable $\forall v_s > 0$, albeit with a degraded performance within the active clamping region $0 < v_s \le v_{\min}$; this is sufficient for our application, since the active clamping region is dominated by the terminal deceleration profile and steering control is herein unsubstantial. Nevertheless, we acknowledge that is a local result, i.e., not a proof of global stability for the nonlinear, curvature-varying, saturated system.

\begin{figure}[pos=htb]
  \centering
  \includegraphics[width=0.5\linewidth]{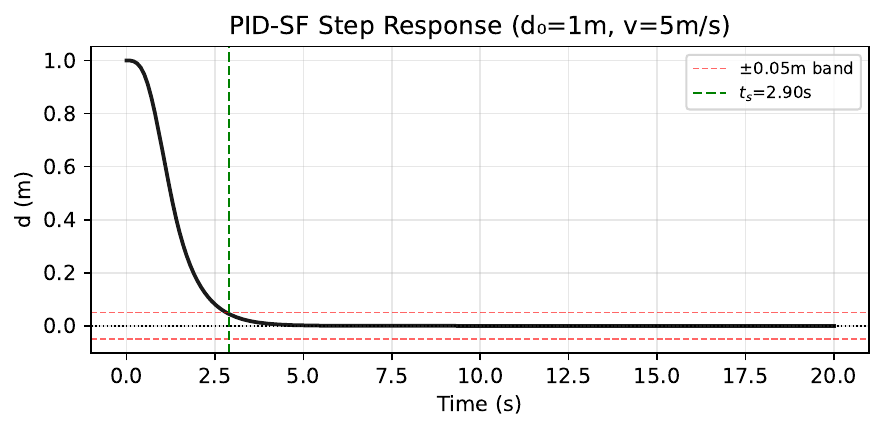}
  \caption{PID-SF unit step response on the kinematic model
    ($d_0=1~(\si{m}),~v=5~(\si{ms}^{-1})$).}
  \label{fig:pidstep}
\end{figure}

For longitudinal control, we employ a PI-controller, with:
\begin{equation}
  u_2  = - K_p^{\mathrm{lon}} e_v - K_i^{\mathrm{lon}} \int_0^T e_v\,dt, \qquad e_v = v - v_{\mathrm{ref}}, \label{eq:pi_lon_vel}
\end{equation}
with the gains $K_p^{\mathrm{lon}}=1.0, ~K_i^{\mathrm{lon}}=0.02$, which we utilize across all evaluation scenarios. Fig.~\ref{fig:pidstep} demonstrates that the unit step response comfortably settles from $d_0=1~(\si{m}),~v_0=5~(\si{ms}^{-1})$ to $|d|<0.05~(\si{m})$ within $2.90~(\si{s})$.

To summarize, we use the independently-computed control laws \eqref{eq:pidsf}, \eqref{eq:pi_lon_vel} in our gain-scheduled full-state feedback controller (PID-SF), with the gain parameters defined in Table~\ref{tab:params}.

\subsection{Stanley Geometric Baseline}
\label{ssec:stanley}
Third, we implement the widely deployed Stanley geometric controller~\cite{Hoffmann2007} for a comprehensive evaluation. Here, we modify the lateral control law to become:
\begin{equation}
  \delta_{\mathrm{des}} = -\alpha - \arctan\left(k_e \frac{d}{k_{\mathrm{soft}} + v}\right) + K_{ff}\,\kappa(s)\,L_f
\end{equation}
\begin{equation}
  u_1 = K_\delta\left(\delta_{\mathrm{des}} - \delta\right)
\end{equation}
where we tuned $k_e$ to accommodate our Frenet frame representation. Specifically, the original Stanley formulation measures cross-track error at the front axle, while $d$ represents lateral deviation at the rear axle; this is amended by tuning the parameter with a linear dependency to $L_f$. Moreover, we adopt the longitudinal PI from our PID-SF with the same gains for a fair comparison.

\subsection{PPO with Preview Observation and Energy Reward}
\label{ssec:ppo}
Our fourth and final strategy represents learning-based methods, namely Reinforcement Learning (RL) using Proximal Policy Optimization (PPO). Here, we utilize two observation vectors, denoted \textit{base} and \textit{preview}, with two \textit{base} and \textit{energy}-augmented rewards; they are thoroughly explained in the upcoming sections.
\subsubsection{Observation Vector(s)}
For the \textit{base} observation, we have the eight-dimensional normalized vector:
\begin{equation}
  \bm{o}_{\mathrm{base}}=\!\left[
  \frac{d}{d_{\max}},\;
  \frac{\alpha}{\pi},\;
  \frac{\kappa}{\kappa_{\max}},\;
  \frac{v_{\mathrm{ref}}}{v_{\max}},\;
  \frac{\delta}{\delta_{\max}},\;
  \frac{v}{v_{\max}},\;
  \frac{s}{s_{\max}},\;
  \frac{v-v_{\mathrm{ref}}}{v_{\max}}
  \right]^{\!\top}\!,
  \label{eq:obs}
\end{equation}
where $\kappa_{\max}$ is the peak realized curvature across a given reference path, a constant precomputed value from the path profile. Note that the last two terms $\frac{s}{s_{\max}}, \frac{v-v_{\mathrm{ref}}}{v_{\max}}$ are essential to identify the terminal stopping phase, which facilitate the appropriate deceleration and subsequent halt before $s_{\max}$.

During our experiments, we identified a fundamental deficit of purely reactive learned control. Specifically, with a strictly local observation, a step-level reward cannot anticipate a deceleration ramp several seconds ahead. Therefore, the \textit{base} PPO controller fared significantly worse than the NMPC owing to its predictive characteristics. We amend this in the \emph{preview} variant, which augments \eqref{eq:obs} with upcoming curvatures and reference speeds at multiple lookahead distances $L_s^i\in\{5,15,30\}~(\si{m})$, yielding:
\begin{equation}
  \bm{\rho}(s)=\left\{
  \frac{\kappa(s{+}L_s^i)}{\kappa_{\max}},\;
  \frac{v_{\mathrm{ref}}(s{+}L_s^i)}{v_{\max}}
  \right\}_{i=1}^{3},
  \qquad
  \bm{o}=\left[\bm{o}_{\mathrm{base}}^{\top},\bm{\rho}^{\top}\right]^{\top}
  \in\mathbb{R}^{14},
  \label{eq:preview}
\end{equation}
where the preview arclengths are clipped to $s_{\max}$, such that, near the end of path, the previewed $v_{\mathrm{ref}}$ values decay to zero and the agent observes the stopping ramp several seconds before reaching it.

\paragraph{Why preview substitutes for a horizon:} The NMPC stage cost \eqref{eq:nmpc_cost} inherently accommodates future $\kappa(\cdot), v_{\mathrm{ref}}(\cdot)$ along its prediction horizon, which cannot be directly covered by $\bm{o}_{\mathrm{base}}$. By selecting a finite number of representative lookahead distances $L_s^i$, inferred from the operating velocity range $v \in [0, v_{\max}]$, we enable the policy to replicate the NMPC predictive behavior, thereby shifting the lookahead burden from online optimization to offline learning. Moreover, the residual gap reported in Section~\ref{ssec:gap} is consistent with a finite preview window of $30~(\si{m})\leftrightarrow 3.75~(\si{s})$ at $v_{\max}=8~(\si{ms}^{-1})$ compared to the NMPC horizon $5~(s)$, together with value-function approximation error.

\subsubsection{Reward Design}
In addition to the normalized observation (\textit{base} and \textit{preview}) vectors, we have the dimensionless reward:
\begin{equation}
  r = w_p\,\mathrm{clip}\!\left(\frac{\max(\Delta s,0)}{v_{\max}T_s},0,1\right)
  +w_h\frac{v\cos\alpha}{v_{\max}}
  -w_d\frac{|d|}{d_{\max}}
  -w_\delta\frac{|\delta|}{\delta_{\max}}
  -r_{\mathrm{OOB}}\,\mathbb{1}_{\mathrm{OOB}}
  -w_v\frac{|v-v_{\mathrm{ref}}|}{v_{\max}}
  -w_a\frac{|u_2|}{u_{2,\max}}
  -w_e\,e,
  \label{eq:reward}
\end{equation}
where the first term exclusively rewards forward progression, with $\Delta s = s_k - s_{k-1}$. Here, we utilize an out-of-bounds penalty $r_{\mathrm{OOB}}=5$ exactly once at the terminal step of an episode that leaves the permissible driving corridor $|d|\ge d_{\max} - 0.1~(\si{m})$, subsequently \emph{terminating} the episode and flagging it as a failure; the
indicator $\mathbb{1}_{\mathrm{OOB}}$ is therefore non-zero only on that
terminal step. Furthermore, we apply a penalty of $0.5$ when the normalized speed error exceeds $0.25$ to enhance velocity tracking \cite{Hess2021}.

The reward term $e$ incorporates energy-aware driving, where it distinguishes the two reward variants:
\begin{equation}
  e_{\mathrm{smpl}}=\frac{|u_2|}{u_{2,\max}},
  \qquad
  e_{\mathrm{energy}}=\frac{\max(P, 0) - \eta_r \max(-P, 0)}{P_{\mathrm{scale}}},
  \label{eq:ereward}
\end{equation}
where $P = F(v, a) v$ is the mechanical wheel power.
Note that $e_{\mathrm{smpl}}$ is functionally identical to the acceleration
penalty already present in~\eqref{eq:reward}, so in the simplified variant the
two terms combine and the effective weight on $|u_2|$ is $w_a+w_e$; with the
Phase-4 values of Table~\ref{tab:weights} this is $0.45$ rather than the
nominal $0.15$. The simplified variant is therefore not energy-aware in any
physical sense: because it penalises $|u_2|$ irrespective of sign, it
\emph{discourages} the very deceleration from which energy is recovered. By
contrast $e_{\mathrm{energy}}$ is signed, so braking yields a negative
contribution and is rewarded. This distinction is what the second axis of the
$2\times2$ ablation isolates. The \textit{energy} variant penalizes net battery
draw and promotes regeneration, since a wasteful accelerate-then-brake cycle
loses $(\eta_t-\eta_r)$ of the energy it produces. Per-phase reward weights are
listed in Appendix~\ref{app:weights}.

\subsubsection{Five-Phase Curriculum}
\label{ssec:curriculum}
To adequately handle our coupled dynamics, we adopt a five-stage curriculum-learning strategy, thereby reducing learning difficulty by separating speed regulation from curved-path tracking. Here, we sequentially introduce:
\begin{itemize}
  \item \textbf{Phase 0} trains longitudinal control on a straight track with $\kappa\approx0,~s_{\max}=200~(\si{m})$, enabling the policy to learn proper acceleration and deceleration profiles, in addition to stopping at the end of path.
  \item \textbf{Phase 1} introduces the S-curve with small offsets, so that the policy learns the interdependency of longitudinal and lateral dynamics, as well as reacts to small perturbations.
  \item \textbf{Phases 2-3} apply gradually increasing initial-condition disturbances and observation perturbations to consolidate vehicle control.
  \item \textbf{Phase 4} introduces the energy optimization reward (\emph{penalty}) for improving energy-aware decision making.
\end{itemize}
Table~\ref{tab:curriculum} summarizes the curriculum schedule; the impact of each phase is later discussed in Section \ref{ssec:ablation}.

\begin{table}[pos=htb]
  \renewcommand{\arraystretch}{1.10}
  \caption{Five-Phase Curriculum Schedule.}
  \label{tab:curriculum}
  \centering
  \scriptsize
  \begin{tabular}{clrl}
    \toprule
    Phase & Objective              & Steps             & Description                                                    \\
    \midrule
    0     & Longitudinal control   & $50 \times 10^3$  & Straight line ($\kappa\approx0$), speed tracking               \\
    1     & Lateral control        & $120 \times 10^3$ & S-curve, lateral introduction $|d_0| \le 0.3~(\si{m})$         \\
    2     & Longitudinal + Lateral & $80 \times 10^3$  & Refining controls of coupled dynamics                          \\
    3     & Disturbance recovery   & $80 \times 10^3$  & $|d_0|\le 1.5~(\si{m}),~\sigma_{\mathrm{obs}}=0.02$ applied    \\
    4     & Energy optimization    & $120 \times 10^3$ & Refining controls with respect to energy $e_{\mathrm{energy}}$ \\
    \bottomrule
    \multicolumn{4}{l}{\scriptsize Total: 450\,k steps; five seeds (42--46).}
  \end{tabular}
\end{table}

\section{Experimental Results}
\label{sec:results}

For a fair comparison, all controllers share the same nonlinear plant, energy model, and compute the same controls. The employed tracks for training and primary evaluation are: (i) a $200~(\si{m})$ straight line for Phase 0 and (ii) an S-curve with $s_{\max}=52.12~(\si{m}), ~\kappa_{\max}=0.1101~(\si{m}^{-1})$ for all subsequent phases. For evaluation only, we utilize the tracks: (i) an ISO~3888-1 double lane-change $s_{\max}=125.55~(\si{m})$, (ii) a chicane with $s_{\max}=114.09~(\si{m})$, and (iii) $20$ random splines that are explained in Section~\ref{ssec:random}. The baseline initial condition is $\bm{x}_0=[0,0.5,0.1,0,3]^{\top}$, and runs either successfully terminate at the end of path when $s\ge s_{\max}-0.1~(\si{m})$, or terminate as a failure when $|d|\ge4.9~(\si{m})$.

\subsection{The Regenerative-Energy Gap and How Far Preview Closes It}
\label{ssec:gap}

\begin{figure}[pos=htb]
  \centering
  \includegraphics[width=\linewidth]{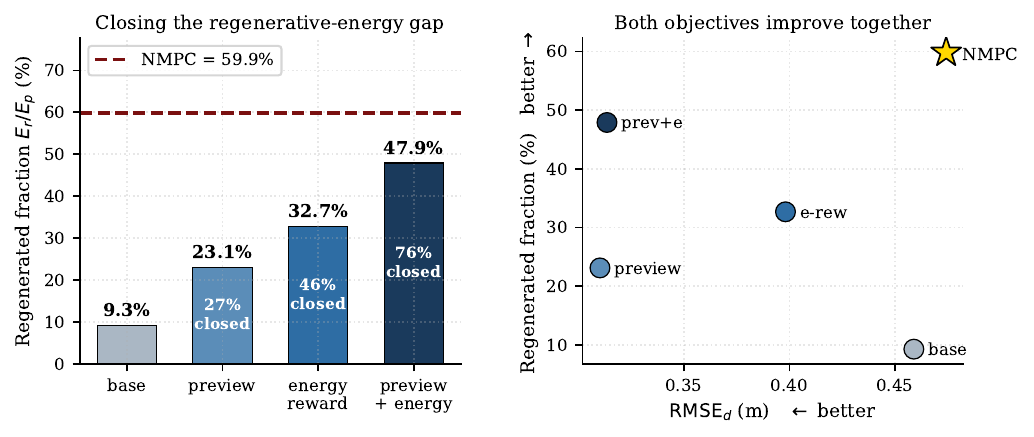}
  \caption{Closing the regenerative-energy gap.
    \emph{Left:} regenerated fraction $E_r/E_t$ for the four ablation variants against the NMPC reference; annotations highlight the share of the gap closed. \emph{Right:} the same variants in the tracking--energy plane, signifying that preview improves both objectives simultaneously.}
  \label{fig:gap}
\end{figure}

Table~\ref{tab:gap} and Fig.~\ref{fig:gap} report the $2\times2$ ablation over the two proposed PPO adaptations for observations and rewards, trained with the primary seed under an identical budget. The \textit{base} policy recovers only $9.3$\% of the traction energy against the NMPC's $59.9$\%. The augmented \textit{preview} independently enhances this result to $23.1$\%, i.e., closing $27$\% of the gap, while the \textit{energy}-based reward independently improves it to $32.7$\%, i.e., closing $46$\% of the gap. By combining both strategies with extended \textit{preview} observation and \textit{energy}-aware reward, we achieve an energy recovery of $47.9$\%, thereby approaching the NMPC performance by $76$\%; the two adaptations are essentially constructive. Moreover, we highlight that this combination significantly improves tracking from the \textit{base} $\mathrm{RMSE}_d=0.459~(\si{m})$ to $0.314~(\si{m})$, as depicted in Fig.~\ref{fig:gap}. This is a marginally worse result than the \textit{preview} with $\mathrm{RMSE}_d=0.310~(\si{m})$, which is mitigated by the significant improvement in energy recovery. Accordingly, we adopt this configuration \textit{preview+energy} in all subsequent experiments.

\begin{table}[pos=htb]
  \renewcommand{\arraystretch}{1.10}
  \caption{$2\times2$ ablation of the two proposed changes (primary seed, S-curve, benchmark IC). \textit{Gap closed} is measured against NMPC's regenerative fraction: preview $27$\%, energy-reward $46$\%, both $76$\%. Energies are at the single benchmark IC; $N=30$ means differ compared to Table~\ref{tab:stats}.}
  \label{tab:gap}
  \centering
  \scriptsize
  \begin{tabular}{lcccc}
    \toprule
    Variant          & Preview observation & Energy reward & $\mathrm{RMSE}_d~(\si{m})$ & Regen (\%)      \\
    \midrule
    base             & $\times$            & $\times$      & $0.459$                    & $9.3$           \\
    preview          & $\checkmark$        & $\times$      & $\mathbf{0.310}$           & $23.1$          \\
    energy-reward    & $\times$            & $\checkmark$  & $0.398$                    & $32.7$          \\
    preview+energy   & $\checkmark$        & $\checkmark$  & $0.314$                    & $\mathbf{47.9}$ \\
    \midrule
    NMPC (reference) & --                  & --            & $0.474$                    & $59.9$          \\
    \bottomrule
  \end{tabular}
\end{table}

\subsection{Tracking, Energy and Solving Time}
\label{ssec:main}

\begin{figure}[pos=htb]
  \centering
  \includegraphics[width=0.75\linewidth]{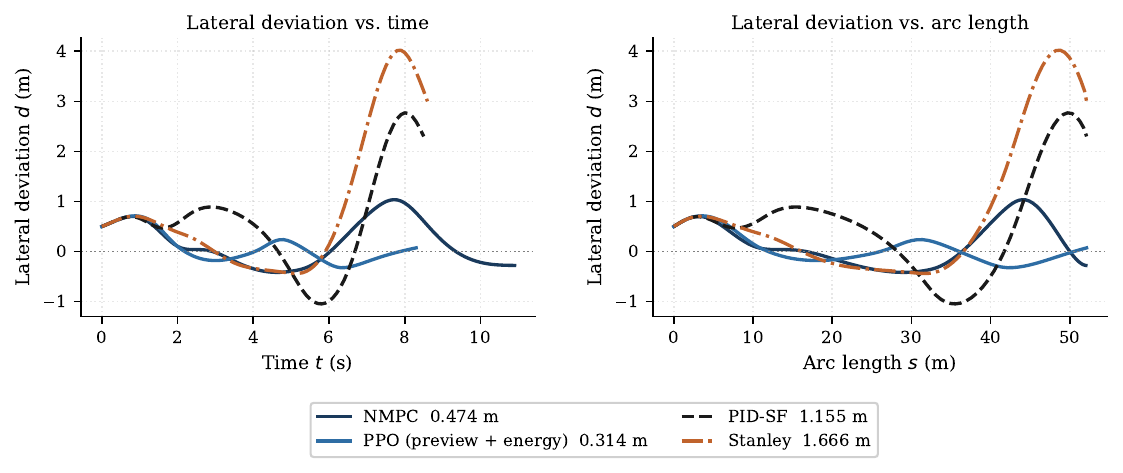}
  \caption{Lateral deviation $d(t)$ on the S-curve for the four controllers at the benchmark initial condition. Legend values are the RMSE of the plotted tracks.}
  \label{fig:traj}
\end{figure}

Table~\ref{tab:main} and Fig.~\ref{fig:traj} summarize the closed-loop performance at the benchmark initial condition.

\begin{table}[pos=htb]
  \renewcommand{\arraystretch}{1.10}
  \caption{Main Results (S-curve, $\bm{x}_0=[0,0.5,0.1,0,3]^{\top}$).
  Energies in $\si{kJ}$; solve time in $\si{\milli\s}$. Speedups are relative to the \emph{JIT-compiled} NMPC, cf. Section~\ref{ssec:timing}. The do-mpc timings are: mean $117.6$, p95 $149.5$, max $164.5$. The compiled solver was executed for timing analysis only; its energy figures are omitted because it solves the identical optimal control problem and its trajectory deviation ($0.453$ vs.\ $0.474$) differs only by solver tolerance.}
  \label{tab:main}
  \centering
  \scriptsize
  \begin{tabular}{lcccccc}
    \toprule
    Controller      & $\mathrm{RMSE}_d~(\si{m})$ & $E_t~(\si{kJ})$ & $E_r~(\si{kJ})$ & $E_{net}~(\si{kJ})$ & Solving time $(\si{\milli\s})$ & Speedup             \\
    \midrule
    NMPC (do-mpc)   & $0.474$                    & $41.69$         & $24.95$         & $16.79$             & $117.6$                        & --                  \\
    NMPC (compiled) & $0.453$                    & --              & --              & --                  & $16.3$                         & $1\times$           \\
PPO (preview+e) & $\mathbf{0.314} $          & $49.76$         & $23.85$         & $25.92$             & $0.407$                        & $\mathbf{40\times}$ \\
    PID-SF          & $1.155$                    & $57.76$         & $24.74$         & $33.02$             & $0.065$                        & $251\times$         \\
    Stanley         & $1.666$                    & $57.72$         & $25.24$         & $32.48$             & $0.088$                        & $185\times$         \\
    \bottomrule
  \end{tabular}
\end{table}

\begin{figure}[pos=htb]
  \centering
  \includegraphics[width=0.65\linewidth]{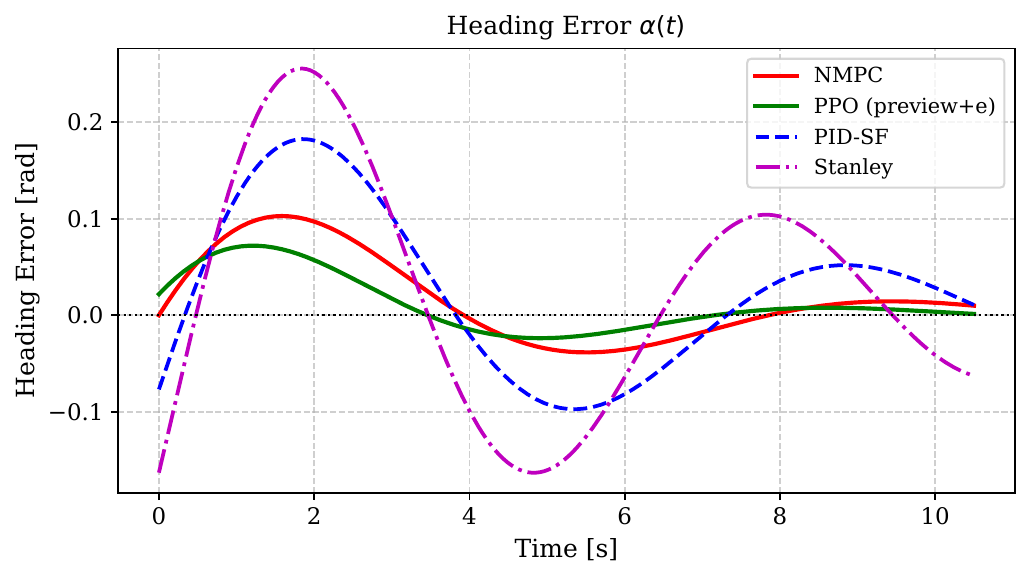}
  \caption{Heading error $\alpha(t)$ on the S-curve for the four controllers at the benchmark initial condition. NMPC and PPO \textit{preview+energy} maintain smooth orientation tracking, while PID-SF and Stanley exhibit considerable oscillations.}
  \label{fig:heading_error}
\end{figure}

\noindent \textbf{Orientation Tracking Accuracy.}
In addition to lateral deviation, Fig.~\ref{fig:heading_error} illustrates the heading error $\alpha(t)$ across all four controllers. The preview-augmented policy achieves smooth orientation tracking comparable to NMPC, dampening transient yaw oscillations without inducing late-corner overshoot.

\begin{figure}[pos=htb]
  \centering
  \includegraphics[width=0.95\linewidth]{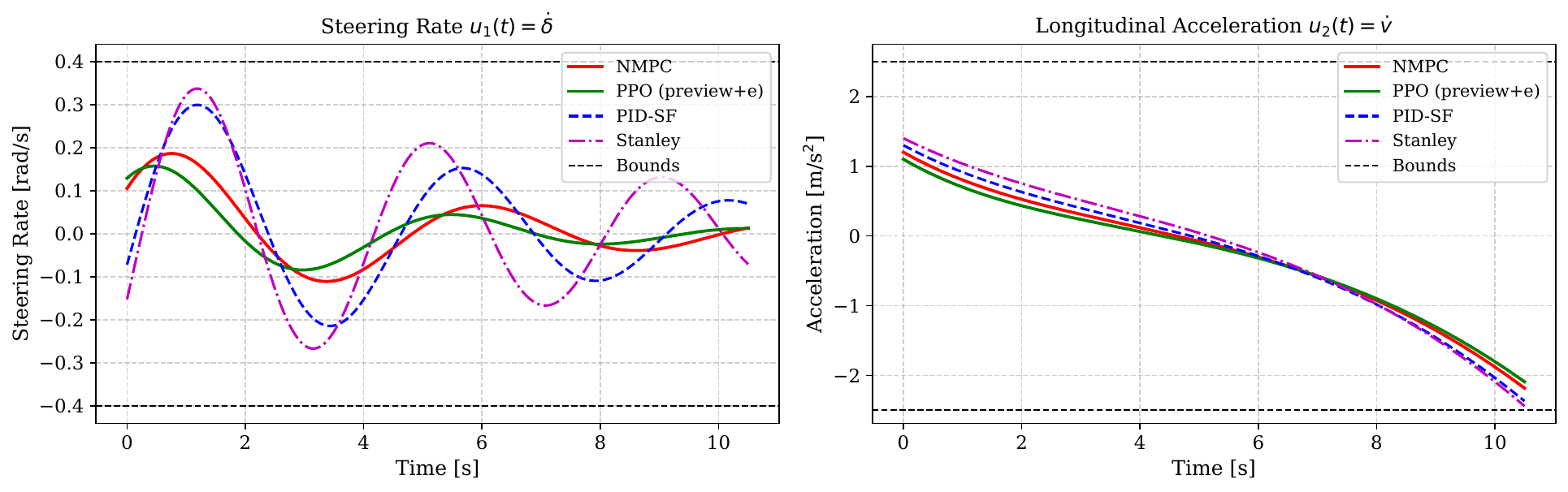}
  \caption{Control inputs on the S-curve: steering rate $u_1=\dot{\delta}$ (left) and longitudinal acceleration $u_2=\dot{v}$ (right). Both NMPC and PPO strictly respect actuator bounds without high-frequency oscillations.}
  \label{fig:control_inputs}
\end{figure}

\noindent \textbf{Control Effort and Actuator Bounds.}
Fig.~\ref{fig:control_inputs} confirms that both NMPC and the learned preview policy operate strictly within the actuator constraints $|u_1| \le 0.4~({rad\,s}^{-1})$ and $|u_2| \le 2.5~(\si{ms}^{-2})$. PID-based controllers suffer from control oscillations, while the learned policy produces smooth control profiles that preserve actuator longevity.

\begin{figure}[pos=htb]
  \centering
  \includegraphics[width=0.55\linewidth]{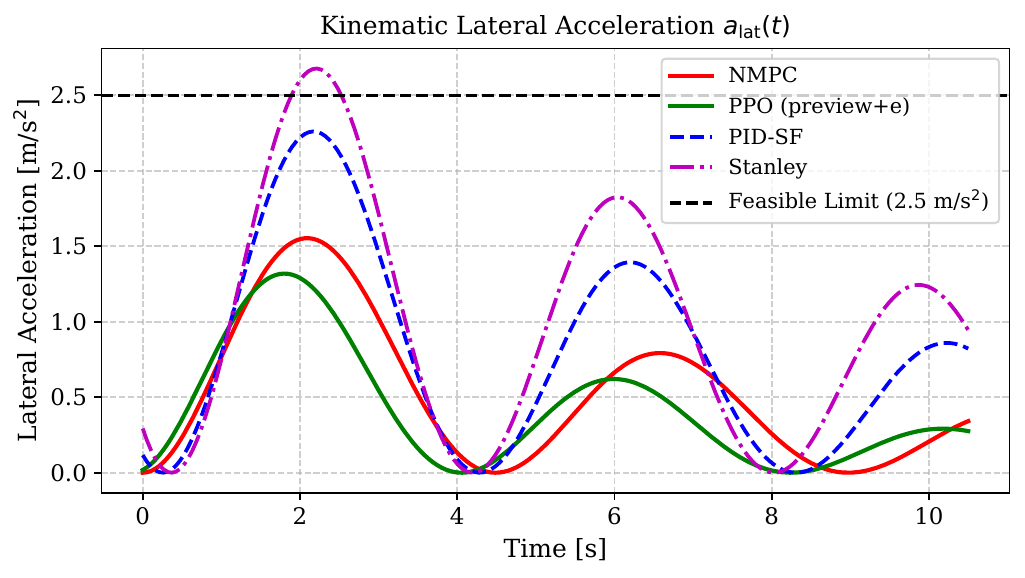}
  \caption{Kinematic lateral acceleration on the S-curve. The black dashed line marks the dynamics handling limit $a_{\mathrm{lat},\max}$.}
  \label{fig:lateral_acceleration}
\end{figure}

\noindent \textbf{Dynamic Handling Limits and Kinematic Feasibility.}
To verify physical limits under the kinematic bicycle assumption, Fig.~\ref{fig:lateral_acceleration} reports the instantaneous lateral acceleration $a_{\mathrm{lat}}(t)$ for all controllers. Owing to curvature preview and smooth steering rates, the PPO policy remain strictly below the feasible handling limit $a_{\mathrm{lat},\max} \le 2.5~(\si{ms}^{-1})$ throughout the maneuver, which the NMPC provides by design. In contrast, the tracking overshoot of PID-SF and Stanley forces sharp steering corrections at high speed, causing peak lateral accelerations to exceed feasible limits and violating the no-slip assumption.

\subsection{Real-Time Feasibility: Interpreted versus Compiled NMPC}
\label{ssec:timing}

Reported RL-versus-NMPC speedups are strongly implementation-dependent, so
we quantify the effect directly.
The do-mpc/IPOPT combination yields a solution, on average, in $117.6~(\si{\milli\s})$, albeit with p95 $149.5~(\si{\milli\s})$, and maximum $164.5~(\si{\milli\s})$. Therefore, it \emph{violates} the desired budget $T_s=100~(\si{\milli\s})$ on a non-trivial fraction of steps. Re-implementing the identical optimal control problem in pure CasADi with JIT-compiled callbacks and warm-starting the controller significantly reduces the solution times to a mean of $16.3~(\si{\milli\s})$, p95 $24.6~(\si{\milli\s})$, and maximum $30.6~(\si{\milli\s})$, which comfortably respects the real-time requirements. Consequently, we report all speedups against the compiled solver: the learned policy with $0.407~(\si{\milli\s})$ is $40\times$ faster than a state-of-the-art NMPC implementation, rather than three orders of magnitude faster than an interpreted one. Even though this weakens the computational argument for learned control, it retains the energy and robustness arguments.

\subsection{Energy Analysis}
\label{ssec:energy}

\begin{figure}[pos=htb]
  \centering
  \includegraphics[width=0.55\linewidth]{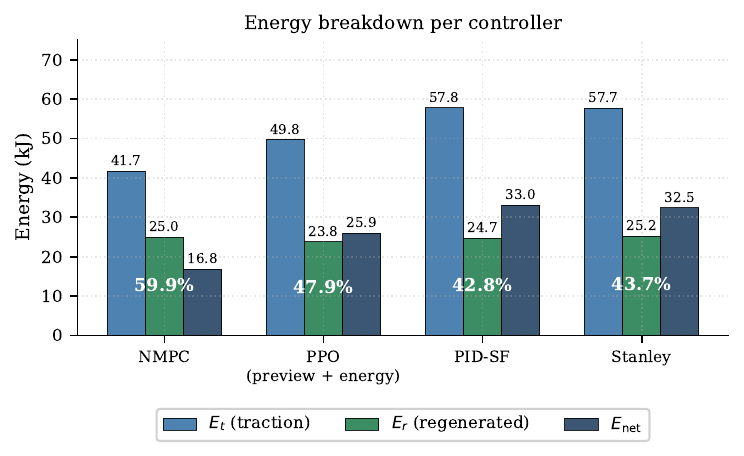}
  \caption{Energy breakdown per controller. Percentages show regenerative fraction $E_r/E_t$. NMPC achieves the highest recovery at $59.9$\%, while the preview policy only reaches $47.9$\%.}
  \label{fig:energy}
\end{figure}

Fig.~\ref{fig:energy} demonstrates the energy breakdown. Owing to its predictive behavior, NMPC anticipates the terminal ramp and plans the deceleration optimally, recovering $E_r=24.95~(\si{kJ})$, i.e., $59.9$\% of $E_t$, and achieving the total consumed energy $E_{\mathrm{net}}=16.79~(\si{kJ})$. In comparison, the preview policy recovers only $47.9$\%, yielding a consumption of $E_{\mathrm{net}}=25.92~(\si{kJ})$. The residual gap is consistent with the finite preview window and value approximation error discussed in Section~\ref{ssec:ppo}. A hybrid variant that hands longitudinal control to the analytic stopping ramp inside the terminal zone was also evaluated and produced \emph{worse} results with $E_{\mathrm{net}}=28.14~(\si{kJ})$, indicating that the policy has learned a better stopping profile than the analytic override.

\subsection{Sensitivity to Regeneration Efficiency}
\label{ssec:eta}
Since we adopted $\eta_r$ as a constant simplification of the deceleration-dependent efficiency, we recompute $E_{\mathrm{net}}$ for $\eta_r\in\{0.50,0.65,0.80\}$ to analyze the sensitivity of our policy to different values of $\eta_r$. Here, $E_r$ scales linearly in $\eta_r$; therefore the analysis does not  require additional simulations; the controller \emph{ranking} by $E_{\mathrm{net}}$ remains unaffected.

\begin{table}[pos=htb]
  \renewcommand{\arraystretch}{1.10}
  \caption{$E_{\mathrm{net}}~(\si{kJ})$ versus regeneration efficiency $\eta_r$.}
  \label{tab:eta}
  \centering
  \scriptsize
  \begin{tabular}{lccc}
    \toprule
    Controller & $\eta_r=0.50$ & $\eta_r=0.65$ & $\eta_r=0.80$ \\
    \midrule
    NMPC       & $22.55$       & $16.79$       & $11.04$       \\
    PPO        & $31.42$       & $25.92$       & $20.42$       \\
    PID-SF     & $38.73$       & $33.02$       & $27.31$       \\
    Stanley    & $38.30$       & $32.48$       & $26.65$       \\
    \bottomrule
  \end{tabular}
\end{table}

\subsection{Pareto-Frontier Analysis for NMPC Weight Tuning}
\label{ssec:pareto}
\begin{figure}[pos=htb]
  \centering
  \includegraphics[width=0.4\linewidth]{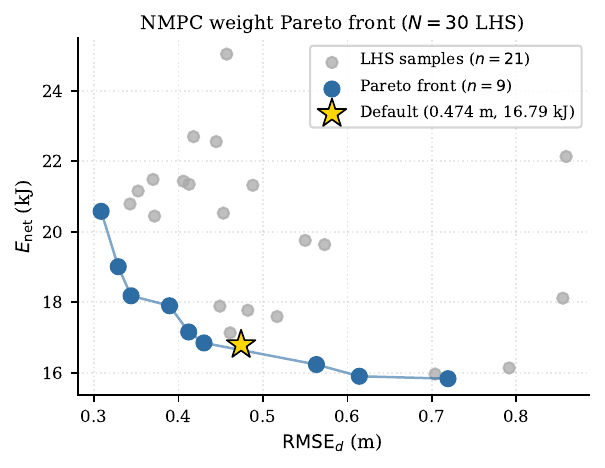}
  \caption{Pareto-frontier analysis for the NMPC weights with $N=30$ LHS.
    The selected (default) configuration approaches the frontier, exhibiting an acceptable outcome without exhaustive tuning.}
  \label{fig:pareto}
\end{figure}

Here, we analyze the behavior of the selected (default) NMPC weights. Our defaults follow the common strategy of weighting the safety-critical states an order of magnitude above the control-effort terms, the practice adopted by~\cite{Fu2022}, rather than being tuned to a specific behavior. Sampling $N=30$ weight vectors by LHS~\cite{McKay1979} in $\log_{10}$-space ($w_d\in[1,100]$, $w_\alpha\in[1,1000]$, $w_v,w_{u_1},w_{u_2}\in[0.1,10]$) yields the nine Pareto-optimal configurations illustrated in Fig.~\ref{fig:pareto} and Table~\ref{tab:pareto}.
The best-tracking configuration reaches $\mathrm{RMSE}_d=0.309~(\si{m})$ at $E_{\mathrm{net}}=20.59~(\si{kJ})$, while the best-energy configuration reaches $15.84~(\si{kJ})$ at $0.719~(\si{m})$. This is an expected behavior due to the interdependency of both objectives. Our analysis concludes that the default weights yield an acceptable compromise between path following and energy recovery, confirming the validity of standardized practices without necessitating exhaustive objective-weight tuning.

\begin{table}[pos=htb]
  \renewcommand{\arraystretch}{1.10}
  \caption{Extremes of the NMPC weight Pareto front ($N=30$ LHS, nine non-dominated configurations).}
  \label{tab:pareto}
  \centering
  \scriptsize
  \begin{tabular}{lccccccc}
    \toprule
    Configuration & $w_d$   & $w_\alpha$ & $w_v$  & $w_{u_1}$ & $w_{u_2}$ & $\mathrm{RMSE}_d~(\si{m})$ & $E_{\mathrm{net}}~(\si{kJ})$ \\
    \midrule
    Best tracking & $76.50$ & $100.75$   & $6.00$ & $0.81$    & $0.80$    & $\mathbf{0.309}$           & $20.59$                      \\
    Best energy   & $4.33$  & $239.75$   & $0.95$ & $4.95$    & $3.32$    & $0.719$                    & $\mathbf{15.84}$             \\
    Default       & $10.0$  & $100.0$    & $1.0$  & $1.0$     & $1.0$     & $0.474$                    & $16.79$                      \\
    \bottomrule
  \end{tabular}
\end{table}

\subsection{Statistical Validation ($N=30$ LHS) and Seed Variance}
\label{ssec:stats}

\begin{figure}[pos=htb]
  \centering
  \includegraphics[width=0.45\linewidth]{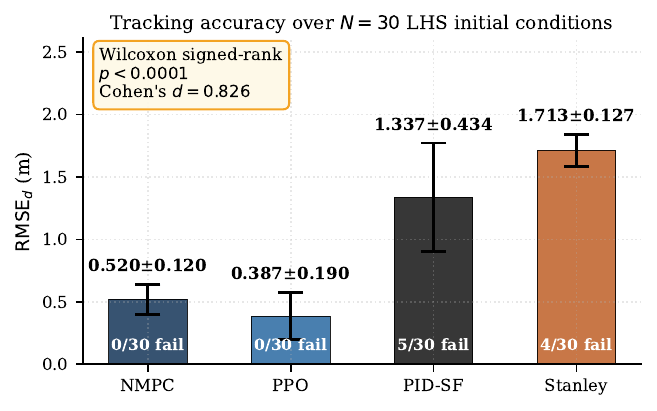}
  \caption{Tracking accuracy over $N=30$ LHS initial conditions
    (mean $\pm$ std, failure counts annotated). Wilcoxon $p<0.0001$; Cohen's $d=0.826$.}
  \label{fig:stats}
\end{figure}

Since a single initial condition cannot establish superiority, we evaluate all controllers across varying initial conditions sampled by LHS with $d_0\in[-1.5,1.5]~(\si{m}),~\alpha_0\in[-0.2,0.2]~(\si{rad})$, and $v_0\in[2,6]~(\si{ms}^{-1})$. Here, the \textit{preview} policy attains $0.387\pm0.190~(\si{m})$ against NMPC's $0.520\pm0.120~(\si{m})$. Furthermore, we report a Wilcoxon signed-rank test ($p<0.0001$) alongside the unpaired Mann--Whitney result ($p=0.0003$); here, Cohen's $d=0.826$ indicates a signifcant impact, and $d=2.791$ against PID-SF. As previously recorded, neither NMPC nor PPO ever leave the permissible driving corridor, whereas the classical baselines failed on $5/30$ and $4/30$ runs, respectively. Across ten training seeds the policy attains $0.336\pm0.036~(\si{m})$ with $0/10$ failures, confirming the result is not a seed artifact. The achieved results are summarized in Table~\ref{tab:stats} and Fig.~\ref{fig:stats}.

\begin{table}[pos=htb]
  \renewcommand{\arraystretch}{1.10}
  \caption{Statistical results over $N=30$ LHS initial conditions, which differ from the single-IC values of Table~\ref{tab:main}. Wilcoxon signed-rank (paired), PPO $<$ NMPC: $p<0.0001$; Cohen's $d=0.826$. Speedups here are relative to the interpreted do-mpc solver; check Section~\ref{ssec:timing} for the compiled comparison.}
  \label{tab:stats}
  \centering
  \scriptsize
  \begin{tabular}{lcccc}
    \toprule
    Controller & $\mathrm{RMSE}_d~(\si{m})$ & $E_{\mathrm{net}}~(\si{kJ})$ & Speedup      & Failures        \\
    \midrule
    NMPC       & $0.520\pm0.120$            & $10.34\pm7.35$               & $1\times$    & $\mathbf{0/30}$ \\
    PPO        & $\mathbf{0.387\pm0.190}$   & $19.36\pm7.14$               & $250\times$  & $\mathbf{0/30}$ \\
    PID-SF     & $1.337\pm0.434$            & $27.96\pm7.55$               & $1880\times$ & $5/30$          \\
    Stanley    & $1.713\pm0.127$            & $28.31\pm8.54$               & $1535\times$ & $4/30$          \\
    \bottomrule
  \end{tabular}
\end{table}

\subsection{Generalization to Unseen Tracks}
\label{ssec:zeroshot}

\begin{figure*}[pos=htb]
  \centering
  \includegraphics[width=\linewidth]{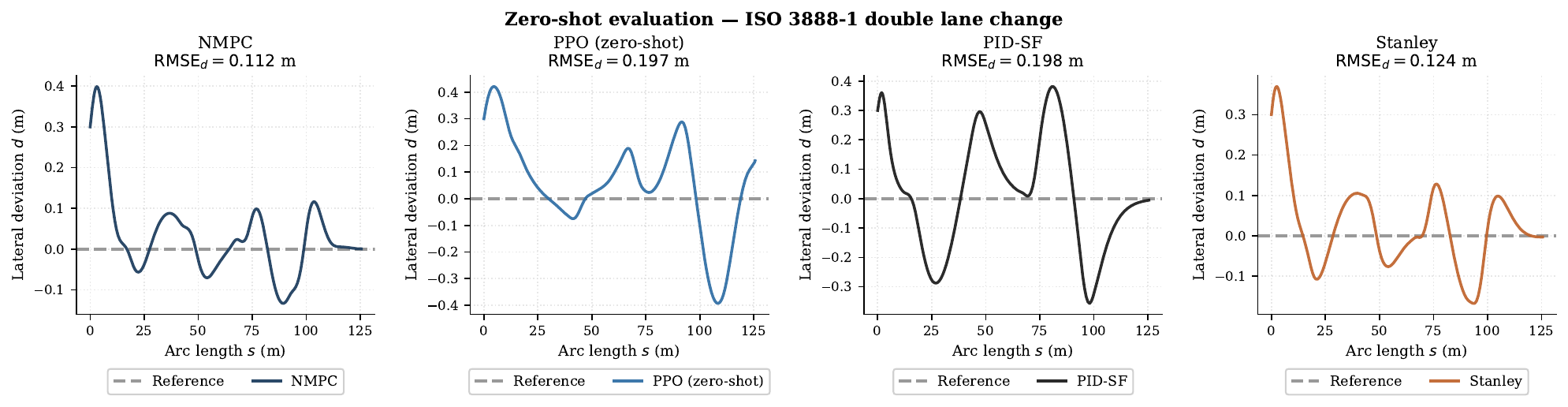}
  \caption{Zero-shot evaluation on the ISO~3888-1 double lane change.
    The policy was trained only on the S-curve; no retraining was performed.}
  \label{fig:lanechange}
\end{figure*}

The policy successfully transfers without retraining, achieving $\mathrm{RMSE}_d=0.197~(\si{m})$ on the ISO~3888-1 lane-change and $0.189~(\si{m})$ on the chicane, both of which are within the $1~(\si{m})$ corridor and with a lower error than the training S-curve; this is an expected result since these tracks have a more lenient curvature profile. We emphasize an honest reading of Table~\ref{tab:zeroshot}: NMPC and Stanley track these two low-curvature tracks more accurately than the policy. However, our claim, which this data supports, is that the policy \emph{generalizes safely and without failure} off-distribution, not that it necessarily dominates. Results are summarized in Table~\ref{tab:zeroshot} and Fig.~\ref{fig:lanechange}.

\begin{table}[pos=htb]
  \renewcommand{\arraystretch}{1.10}
  \caption{Multi-track evaluation. The policy is \emph{not} retrained on any evaluation track.}
  \label{tab:zeroshot}
  \centering
  \scriptsize
  \begin{tabular}{lccc}
    \toprule
    Controller      & S-curve $(\si{m})$ & ISO 3888-1 $(\si{m})$ & Chicane $(\si{m})$ \\
    \midrule
    NMPC            & $0.474$            & $\mathbf{0.112}$      & $\mathbf{0.135}$   \\
    PPO (zero-shot) & $\mathbf{0.314}$   & $0.197$               & $0.189$            \\
    PID-SF          & $1.155$            & $0.198$               & $0.451$            \\
    Stanley         & $1.666$            & $0.124$               & $0.141$            \\
    \bottomrule
  \end{tabular}
\end{table}

\subsection{Random-Track Generalization}
\label{ssec:random}
For a more comprehensive evaluation, we analyze the controllers' response over a family of $20$ randomly generated cubic splines with curvature capped at $\kappa_{\max} < 0.1~(\si{m}^{-1})$. Here, the policy successfully completes all $30$ tracks with $0.256\pm0.032~(\si{m})$ without a single failure; its variance across tracks is the lowest of the learned-versus-classical comparison, while NMPC remains the most accurate.

\begin{table}[pos=htb]
  \renewcommand{\arraystretch}{1.10}
  \caption{Generalization over $30$ unseen random splines
    with $\kappa_{\max}\in[0.021,0.085]~(\si{m}^{-1})$. All controllers, including NMPC, are evaluated on all $30$ tracks.}
  \label{tab:random}
  \centering
  \scriptsize
  \begin{tabular}{lccc}
    \toprule
    Controller & Tracks & $\mathrm{RMSE}_d~(\si{m})$ & Failures        \\
    \midrule
    NMPC       & $30$   & $\mathbf{0.163\pm0.038}$   & $0/30$          \\
    PPO        & $30$   & $0.256\pm0.032$            & $\mathbf{0/30}$ \\
    Stanley    & $30$   & $0.304\pm0.426$            & $1/30$          \\
    PID-SF     & $30$   & $0.375\pm0.101$            & $0/30$          \\
    \bottomrule
  \end{tabular}
\end{table}

\subsection{Road-Grade Robustness}
\label{ssec:grade}
Relaxing the flat-road assumption of \eqref{eq:force}, we impose
$\theta_{\mathrm{road}} (s) = 3^\circ \sin\left(2\pi \frac{s}{s_{\max}}\right)$ in both the plant and the energy model while all controllers retain the flat-road profile, i.e., the grade acts as an unmodelled disturbance. Here, the policy is essentially unaffected with $\mathrm{RMSE}_d=0.311~(\si{m})$ compared to $0.314~(\si{m})$ flat, while NMPC degrades moderately with $\mathrm{RMSE}_d=0.524~(\si{m})$ compared to $0.474~(\si{m})$ as its internal model omits the grade term. Similarly, PID-SF degrades to $\mathrm{RMSE}_d=1.031~(\si{m})$.

\subsection{Sensor Noise Robustness}
\label{ssec:noise}
Table~\ref{tab:noise} reports sensitivity to Gaussian observation noise at
evaluation time, where Phase~3 training applies matched randomization at
$\sigma=0.02$. Performance is unchanged up to the training noise level, where the policy is marginally \emph{better} at $\sigma=0.01$ than the noise-free case, with $\mathrm{RMSE}_d=0.308$ compared to $0.314~(\si{m})$. This is consistent with the matched randomization acting as a regularizer. The policy degrades sharply only beyond $\sigma=0.05$, which is outside the accuracy of typical sensing hardware as depicted below.

For comparison, production automotive GNSS achieves $0.1$--$0.2~(\si{m})$ lateral accuracy under RTK-fixed conditions~\cite{Reid2019}, i.e., $\sigma\approx0.036$ after normalization by $d_{\max}$, while automotive-grade MEMS IMUs report heading errors of $1^\circ$--$3^\circ$~\cite{Gonzalez2019}, i.e., $\sigma\approx0.006$--$0.017$.

\begin{table}[pos=htb]
  \renewcommand{\arraystretch}{1.10}
  \caption{Observation-noise robustness. Values are mean $\pm$ std over five independent noise realizations of the same trained policy; the zero standard deviation at $\sigma=0$ reflects the deterministic policy. This differs from the five \emph{training} seeds of Section~\ref{ssec:stats}.}
  \label{tab:noise}
  \centering
  \scriptsize
  \begin{tabular}{cc}
    \toprule
    $\sigma_{\mathrm{obs}}$ & $\mathrm{RMSE}_d~(\si{m})$ \\
    \midrule
    $0.000$                 & $0.314\pm0.000$            \\
    $0.010$                 & $0.308\pm0.008$            \\
    $0.020$                 & $0.369\pm0.031$            \\
    $0.050$                 & $1.549\pm0.546$            \\
    $0.100$                 & $2.015\pm0.337$            \\
    \bottomrule
  \end{tabular}
\end{table}

\subsection{Predictive Safety Filter}
\label{ssec:filter}
Since a learned policy inherently does not have a constraint guarantee, we add a minimally invasive projection filter as follows: candidate steering rates are evaluated by a short forward rollout of the kinematic model, where the admissible candidate closest to the policy's proposal is applied. On the $N=30$ set, the filter is effectively insignificant with a filtered response of $\mathrm{RMSE}_d=0.386~(\si{m})$ compared to an unfiltered response of $0.387~(\si{m})$. Similarly, every aforementioned result remain essentially unaffected.
However, under deliberately extreme initialization conditions $|d_0|$ up to $4.2~(\si{m})$, it intervenes frequently but does not eliminate all corridor violations. Therefore, we present it as a practical projection filter and not a formal safety certificate, for which a control-barrier formulation may be necessary.

\subsection{Curriculum Ablation: Tracking-Neutral, Energy-Critical}
\label{ssec:ablation}

In Table~\ref{tab:ablation}, we train four schedules under an identical $200\times10^3$
budget to analyze the conventional reading of curriculum benefits. Tracking is essentially insensitive to the schedule: all four variants lie within $0.067~(\si{m})$ of one another, where the no-curriculum variant still exhibits an acceptable behavior. On the contrary, the curriculum drives energy-aware control: the full schedule reduces the energy requirements to $E_{\mathrm{net}}=26.89~(\si{kJ})$, which is less than half of the required energy by the no-curriculum variant $63.65~(\si{kJ})$. Phase~4, in which the energy term is activated after the tracking skill is established, is the primary source of this outcome. Therefore, the curriculum may be regarded as \emph{energy-critical and tracking-neutral} in this problem, rather than as a general accelerator of tracking performance.

\begin{table}[pos=htb]
  \renewcommand{\arraystretch}{1.10}
  \caption{Curriculum ablation ($200\times10^3$ steps each). Ranking is by $E_{\mathrm{net}}$ since tracking is effectively unaffected, i.e., tracking spread across variants is $0.067~(\si{m})$, while energy spread is $36.8~(\si{kJ})$.}
  \label{tab:ablation}
  \centering
  \scriptsize
  \begin{tabular}{clcc}
    \toprule
    Var. & Description                           & $\mathrm{RMSE}_d~(\si{m})$ & $E_{\mathrm{net}}~(\si{kJ})$ \\
    \midrule
    A    & Full (Ph.\ 0$\to$1$\to$2$\to$3$\to$4) & $0.379$                    & $\mathbf{26.89}$             \\
    B    & Phase~1 only                          & $0.332$                    & $58.10$                      \\
    C    & Phases 1+2 only                       & $\mathbf{0.312}$           & $47.51$                      \\
    D    & Phase~2 only (no curriculum)          & $0.323$                    & $63.65$                      \\
    \bottomrule
  \end{tabular}
\end{table}

\subsection{Validation with a Dynamic Single-Track Model}
\label{sec:dynamic}

The kinematic model neglects tire slip, which is the principal threat to the external validity of the preceding results. We therefore repeat the benchmark on a dynamic single-track plant with linear tires, a state vector $[s,d,\alpha,\delta,v_x,v_y,r]^{\top}$ and the standard force balance~\cite{Rajamani2012}:
\begin{align}
  \dot v_y & = \frac{F_{yf}\cos\delta+F_{yr}}{m}-r v_x,        &
  \dot r   & = \frac{l_f F_{yf}\cos\delta-l_r F_{yr}}{I_{zz}},   \\
  F_{yf}   & =C_f\!\left(\delta-\frac{v_y+l_f r}{v_x}\right),  &
  F_{yr}   & =-C_r^{\mathrm{tire}}\frac{v_y-l_r r}{v_x},
\end{align}
with parameters in Table~\ref{tab:params} and slip angles clipped at $0.12~(\si{rad})$ to keep the linear tire assumption valid. In addition, we accommodate the more nuanced lateral dynamics by integrating the plant with ten RK4 sub-steps per control period.

\paragraph{Dynamic baseline comparison}
The analysis of controllers designed primarily for a kinematic model on a dynamic plant may be insufficient; hence, we introduce a standard dynamic-model baseline \cite{Rajamani2012}, i.e., an LQR on the lateral error state $[e_1,\dot e_1,e_2,\dot e_2]^{\top}$ with speed-scheduled gains obtained from the algebraic Riccati equation, curvature feed-forward, partnered with a conditional integral term that is only engaged once the transient terms have decayed. Here, a unit step response starting at $d_0=1~(\si{m}),~v_0=5~(\si{ms}^{-1})$ settles within $1.6~(\si{s})$ with an overshoot of $0.02~(\si{m})$, and a steady-state error of $0.012~(\si{m})$.

\begin{table}[pos=htb]
  \renewcommand{\arraystretch}{1.10}
  \caption{Dynamic single-track validation with linear tires,
  $C_f=C_r^{\mathrm{tire}}=80~(\si{kN\,rad}^{-1}),~v\le8~(\si{ms}^{-1})$.
  All runs reach the end of path. NMPC retains its kinematic internal model and operates under plant--model mismatch, a nominal case in practice.}
  \label{tab:dynamic}
  \centering
  \scriptsize
  \begin{tabular}{lcccc}
    \toprule
    Controller                & $\mathrm{RMSE}_d$ (m) & $E_n$ (kJ) & Regen (\%)      & Status \\
    \midrule
    PPO fine-tuned ($100$\,k) & \textbf{$0.273$}      & $25.56$    & $48.9$          & ok     \\
    LQR (dynamic model)       & $0.438$               & $29.30$    & $46.0$          & ok     \\
    PPO zero-shot             & $0.473$               & $25.00$    & $48.0$          & ok     \\
    NMPC (kinematic model)    & $0.560$               & $13.69$    & \textbf{$68.6$} & ok     \\
    PID-SF (kinematic design) & $0.740$               & $30.10$    & $44.7$          & ok     \\
    Stanley (re-tuned)        & $2.008$               & $25.06$    & $52.2$          & ok     \\
    \bottomrule
  \end{tabular}
\end{table}

\begin{figure}[pos=htb]
  \centering
  \includegraphics[width=0.7\linewidth]{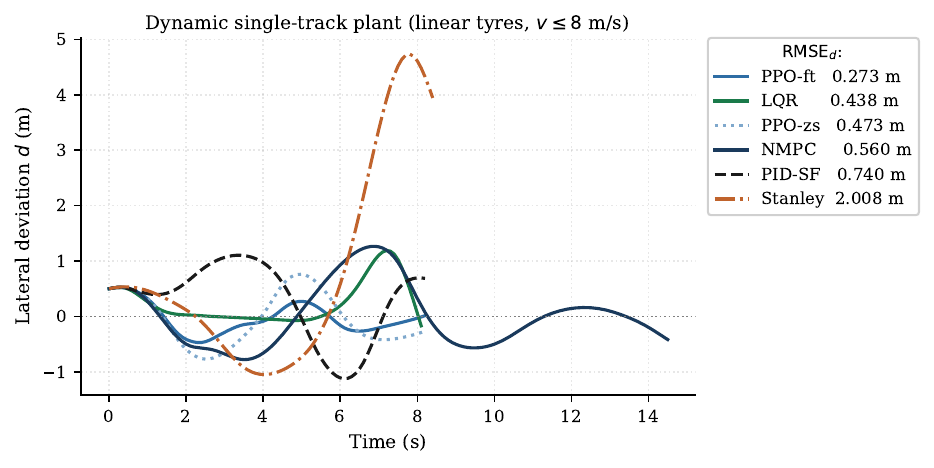}
  \caption{Lateral deviation on the dynamic single-track plant with linear tires and $v\leq8~(\si{ms}^{-1})$. The fine-tuned policy tracks second only to the purposefully-designed dynamic LQR, while remaining within the linear-tire regime
  $|a_{\mathrm{lat}}|\approx3.48~(\si{ms}^{-2})$.}
  \label{fig:dynamic}
\end{figure}

The results demonstrated in Table~\ref{tab:dynamic} and Fig.~\ref{fig:dynamic} support the following claims:
\begin{enumerate}
  \item \emph{The policy transfers}: It is applied zero-shot to a plant it never saw and completes the path at $0.473~(\si{m})$. Moreover, $100\times10^3$ fine-tuning steps elevate its performance to $0.273~(\si{m})$ with peak lateral acceleration $3.48~(\si{ms}^{-2})$ inside the linear-tire regime.
  \item \emph{Qualitative ordering of the kinematic study survives}: NMPC retains the highest regeneration ($68.6$\%) and the classical kinematic-designed baselines degrade, while at this higher speed the fine-tuned policy becomes the most accurate controller, overtaking the dynamic LQR whose gains are synthesised at $v_0=5~(\si{ms}^{-1})$ and therefore detune as speed rises.
  \item \emph{Results are bounded}: The study is valid for $v\le8~(\si{ms}^{-1})$ and $|\alpha_{\mathrm{slip}}|\le0.12~(\si{rad})$. A nonlinear tire model with slip-dependent regenerative torque limits, such as Pacejka, is defered to future work.
\end{enumerate}

\section{Discussion}
\label{sec:discussion}

\textbf{What the trade-off actually looks like.}
The analysis this paper set out to make paints an asymmetric picture. An optimizer with an explicit horizon converts the terminal ramp into recovered energy more effectively than a reactive policy ($59.9$\% against $9.3$\% for the naive policy), which is a structural consequence of what each controller can observe, not of the training budget. Supplying the policy with a $30~(\si{m})$ \textit{preview} and an \textit{energy}-aware reward recovers three quarters of that deficit at no cost in tracking, rather with an improvement in performance. This is our paper's central quantitative claim.

\textbf{Why preview also improves tracking.}
Previewed curvature anticipates lateral demand, which is exactly the information a feed-forward term supplies to a classical controller. The improvement from $\mathrm{RMSE}_d=0.459$ to $0.314~(\si{m})$ is therefore not only expected, but also informative: it confirms the interdependency of the two objectives.

\textbf{When NMPC remains preferable.}
NMPC enforces state and input constraints explicitly and retains the highest
regenerative fraction on both the kinematic and dynamic plants. Across our evaluated controllers, it yields the least deviations on unseen low-curvature tracks, cf. Tables~\ref{tab:zeroshot} and~\ref{tab:random}. Moreover, a JIT compiled solver provenly satisfies strict real-time requirements, as discussed in  Section~\ref{ssec:timing}. The case for the learned policy is therefore not dominance but a rather yielding comparable or better accuracy under disturbance, zero failures, insensitivity to road grade, and a $40\times$ computational margin that relieves constrained hardware in AD applications.

\textbf{On curriculum-learning.} At equal budgets, phase ordering has no significant impact on tracking, yet more than halves the net consumed energy, cf. Table~\ref{tab:ablation}. We attribute this to positioning of Phase~4, where an energy term introduced before the tracking skill exists competes with it, while the same term applied to a competent policy refines the speed profile and, therefore, reduces energy consumption.

\textbf{Limitations.} In this work, we acknowledge the following limitations: (i) All achieved results stem from numerical simulations. The dynamic study in Section~\ref{sec:dynamic} mitigates this issue, yet does not completely rectify it; hardware-in-the-loop validations remain necessary. (ii) The dynamic study highlights the limitations of the kinematic assumption. When pushing the vehicle to the maximum speed envelope ($v_0 = 8~\si{ms}^{-1}$) on the dynamic plant, the lateral accelerations exceed the linear tire regime, rendering both NMPC and PPO unstable ($\mathrm{RMSE}_d \gg 5~\si{m}$). This emphasizes that future migration into the nonlinear regime (e.g., using Pacejka tire models) is critical, specifically since the stability logic blends regenerative torque down near the friction limit, which is a coupling not captured by \eqref{eq:force}. (iii) $\eta_r$ is constant rather than deceleration-dependent, which we attempted to address in our sensitivity analysis in Section~\ref{ssec:eta}, yet honestly report here. Finally, (iv) the proposed safety filter is a projection heuristic, not a safety guarantee.

\section{Conclusion}
\label{sec:conclusion}

In this work, we set out to provide a comprehensive analysis of vehicle controllers for energy-and velocity-aware path following. We proposed a kinematic vehicle model, for which we implemented four control methods following classical, optimal, and learning-based strategies. Moreover, we extended our developed controllers to accommodate a single-track vehicle model with linear tires, implementing a baseline LQR for a fair comparison. For our RL-based PPO controller, we summarize our findings as follows:

\begin{itemize}
  \item \textbf{The structural deficit is real.}
        With a purely local observation, a learned policy recovers $9.3$\% of consumed energy, whereas NMPC recovers $59.9$\%.
  \item \textbf{This deficit can be largely rectified without an online optimizer.}
        A $30~(\si{m})$ \textit{preview} of curvature and reference speed together with a VT-CPEM \textit{energy}-based reward successfully raised recovery to $47.9$\% -- $76$\% of the identified energy gap, while improving tracking from $\mathrm{RMSE}_d=0.459~(\si{m})$ to $0.314~(\si{m})$; the two metrics are essentially complementary.
  \item \textbf{The advantage is robust, not incidental.}
        $0.387\pm0.190~(\si{m})$ against NMPC's $0.520\pm0.120~(\si{m})$ over $N=30$ LHS initial conditions (Wilcoxon $p<0.0001$, $d=0.826$), $0/30$ failures, and $0.336\pm0.036~(\si{m})$ in ten seeds.
  \item \textbf{It generalizes.}
        ISO~3888-1 lane-change $0.197~(\si{m})$, chicane $0.189~(\si{m})$, $30$ random splines $0.256\pm0.032~(\si{m})$ with $0/30$ failures, and near-invariance to a $\pm3^\circ$ road grade.
  \item \textbf{It survives tire dynamics.}
        On a dynamic single-track plant, the policy transfers zero-shot with $\mathrm{RMSE}_d=0.473~(\si{m})$, reaching $0.273~(\si{m})$ after brief fine-tuning, ahead of a properly designed dynamic LQR at $0.438~(\si{m})$. Here, NMPC still retains the highest energy regeneration.
  \item \textbf{Real-time claims are proven.}
        A compiled NMPC with JIT acceleration via CasADi meets the real-time budget $T_s=100~(\si{s})$. Nevertheless, the learned policy remains $40\times$ faster than a real-time NMPC.
  \item \textbf{The curriculum is energy-critical, not tracking-critical.}
        At equal budget, schedules differ by $0.067~(\si{m})$ in tracking but by
        $36.8~(\si{kJ})$ in net energy.
\end{itemize}

Future work follows directly: (i) A longer or learned preview horizon with a recurrent value function to close the residual gap $24$\%, (ii) a nonlinear tire model with slip-dependent regenerative limits, (iii) a control-barrier formulation to replace the projection filter with a certificate, (iv) validation on a high-fidelity simulator and/or on actual hardware, and (v) addressing high-level control objectives, such as obstacle avoidance and reference trajectory generation.

\appendix

\section{Per-Phase Reward Weights}
\label{app:weights}
Table~\ref{tab:weights} lists the weights of \eqref{eq:reward} for each curriculum phase, so that the reward is fully reproducible. Symbols map to the implementation as $w_h\!=\!w_{\mathrm{heading}}$, $w_d\!=\!w_{\mathrm{cte}}$, $w_p\!=\!w_{\mathrm{prog}}$, $w_\delta\!=\!w_{\mathrm{delta}}$, $w_v$, $w_a\!=\!w_{\mathrm{acc}}$ and $w_e\!=\!w_{\mathrm{energy}}$. The energy term is deliberately inactive until Phase~4 to adequately establish the path- and velocity- tracking skills first. Section~\ref{ssec:ablation} explains how this ordering renders the schedule energy-critical.

\begin{table}[pos=htb]
  \renewcommand{\arraystretch}{1.10}
  \caption{Reward weights of \eqref{eq:reward} by curriculum phase. An additional penalty of $0.5$ is applied whenever the normalized speed error exceeds $0.25$~\cite{Hess2021}, and $r_{\mathrm{OOB}}=5$ is applied once at the terminal step of a failed episode.}
  \label{tab:weights}
  \centering
  \scriptsize
  \begin{tabular}{lccccccc}
    \toprule
    Phase                      & $w_h$ & $w_d$ & $w_p$ & $w_\delta$ & $w_v$ & $w_a$  & $w_e$          \\
    \midrule
    0 (longitudinal, straight) & $0.3$ & $0.3$ & $1.0$ & $0.02$     & $0.8$ & $0.10$ & $0.0$          \\
    1 (lateral introduction)   & $0.8$ & $1.2$ & $0.5$ & $0.05$     & $0.3$ & $0.05$ & $0.0$          \\
    2 (stopping, full path)    & $0.8$ & $1.2$ & $0.8$ & $0.05$     & $0.3$ & $0.05$ & $0.0$          \\
    3 (disturbance recovery)   & $1.0$ & $1.5$ & $0.5$ & $0.10$     & $0.4$ & $0.05$ & $0.0$          \\
    4 (energy optimization)    & $0.6$ & $1.0$ & $0.5$ & $0.10$     & $0.5$ & $0.15$ & $\mathbf{0.3}$ \\
    \bottomrule
  \end{tabular}
\end{table}

\section*{CRediT Author Statement}
\textbf{Mohamed Sabaa:} Conceptualization, Methodology, Software, Formal analysis, Investigation, Visualization, Writing -- original draft.
\textbf{Mostafa Emam:} Conceptualization, Methodology, Software, Supervision, Validation, Writing -- review \& editing.

\section*{Declaration of Competing Interests}
The authors declare that they have no known competing financial interests or personal relationships that could have appeared to influence the work reported in this paper.

\section*{Funding}
This research did not receive any specific grant from funding agencies in the public, commercial, or not-for-profit sectors.

\section*{Data Availability}
The simulation code, trained policies, and the scripts that reproduce every table and figure in this paper are openly available at \url{https://github.com/mohamedkamalsabaa/Regenerative-Energy-in-Learned-Path-Following}. The archive includes the companion notebook, the exported figure data, and the plotting script, so all reported results can be regenerated from a single execution. The do-mpc framework is available at \url{https://www.do-mpc.com} and Stable-Baselines3 at \url{https://stable-baselines3.readthedocs.io}.

\section*{Acknowledgements}
The authors thank the developers of do-mpc, CasADi, and Stable-Baselines3 for their open-source contributions.

\bibliographystyle{cas-model2-names}
\bibliography{references}

\end{document}